%% file: arxiv.tex
\pdfoutput=1
\documentclass{arxiv2026}

\makeatletter
\renewcommand*{\@titlefoot}{\scriptsize Preprint.\hfill}
\makeatother

\usepackage{times}
\usepackage{microtype}
\usepackage{booktabs}
\usepackage{mathtools}
\usepackage{array}
\usepackage{enumitem}
\usepackage{mdframed}
\usepackage{needspace}

\graphicspath{{figures/}{./}}
\hypersetup{
  colorlinks=true,
  linkcolor=blue!45!black,
  citecolor=blue!45!black,
  urlcolor=blue!45!black,
  pdftitle={The Label Complexity of Useful Class-Conditional Prediction Sets under Distribution Shift},
  pdfauthor={Weijia Han, Lisha Qu, Tianxin Zhou, Zhenda Li, Liying Liang}
}
\newcommand{\E}{\mathbb{E}}

\newcommand{\cvar}{\mathrm{CVaR}_{10}}
\providecommand{\qed}{\hfill\ensuremath{\square}}
\newcolumntype{L}[1]{>{\raggedright\arraybackslash}p{#1}}

\mdfdefinestyle{altkeyresult}{
  linecolor=black!55,
  linewidth=0.55pt,
  backgroundcolor=black!2,
  innertopmargin=5pt,
  innerbottommargin=5pt,
  innerleftmargin=7pt,
  innerrightmargin=7pt,
  skipabove=7pt,
  skipbelow=7pt,
  nobreak=true
}

\title[Useful Class-Conditional Prediction Sets]{The Label Complexity of Useful Class-Conditional Prediction Sets under Distribution Shift}

\author[Han et al.]{%
  \Name{Weijia Han},
  \Name{Lisha Qu},
  \Name{Tianxin Zhou},
  \Name{Zhenda Li}, \and
  \Name{Liying Liang}
}

\begin{document}
\maketitle

\begin{abstract}
\input{sections/alt_abstract}
\end{abstract}


\input{sections/alt_intro}
\input{sections/alt_related}
\input{sections/alt_problem}
\input{sections/alt_identification}
\input{sections/alt_audit}
\input{sections/alt_experiments}
\input{sections/alt_discussion}

\Needspace{6\baselineskip}
\section*{AI Disclosure}
In this work, we used generative AI tools only to assist with implementing methods in software,
editing code, and polishing the manuscript for clarity and readability. We did not use these tools
to generate, clean, or reformat datasets; develop theoretical models or mathematical claims;
formulate hypotheses; design or evaluate research methodology or experiments; write proofs; or
interpret results. Translation and qualitative or thematic analysis were not applicable. The
authors reviewed every change produced with this assistance, tested the resulting code, and take
responsibility for the final paper and accompanying artifacts.

\bibliography{refs}

\appendix
\input{sections/alt_proofs}
\input{sections/alt_appendix_experiments}

\end{document}

%% file: sections/alt_abstract.tex
Prediction sets can make deployed classifiers safer by returning several plausible labels when a
single prediction is uncertain. Their value depends on classwise reliability: average coverage can
meet its target while rare or difficult classes fail repeatedly. This concern is sharper after
distribution shift, when calibration labels come from a source environment but reliability is
needed on the target. We ask what labeled source data and unlabeled target inputs reveal about
class-conditional prediction sets, and when target labels are necessary. Under unrestricted
joint shift, two target laws can produce the same observable data while requiring different
classwise thresholds; any label-free rule covering both must enlarge its sets on one law. We give
a labeled target audit that estimates the missing quantiles with a simultaneous guarantee.
Probability-scale error is invariant to increasing score transformations, and threshold recovery
follows under local regularity. At fixed confidence, achieving threshold tolerance $\varepsilon$
with fixed, nonadaptive class-stratified labeled pairs has total complexity
$\Theta(K\varepsilon^{-2}\log K)$, or $\Theta(\varepsilon^{-2}\log K)$ labels per class under
equal allocation. Class imbalance creates
a separate acquisition cost; for foreground class probabilities of order $1/K$, the mixed-stream
label complexity is also $\Theta(K\varepsilon^{-2}\log K)$ at fixed confidence. Experiments
on action-recognition and image shifts show that marginal coverage can conceal severe class
failures and that source classwise calibration depends on the shift. The results connect the
information available at deployment to the target labels needed for useful class-conditional
prediction.

%% file: sections/alt_intro.tex
\section{Introduction}
\label{sec:intro}

Prediction sets express uncertainty by returning several plausible labels when a classifier is
unsure. Under exchangeability, conformal calibration gives a finite-sample coverage guarantee
\citep{vovk2005,lei2018}. Yet marginal coverage averages over the target population, allowing
common and easy classes to hide repeated failures on uncommon or difficult ones.
Class-conditional coverage instead requires the target level separately for every class.

This distinction matters when errors have different costs. In medical triage or activity
monitoring, a missed uncommon class may matter even when average accuracy and coverage are high.
A classwise guarantee makes that failure visible. It also exposes the tension between reliability
and usefulness: returning every label is safe but gives little guidance, so prediction sets
should not be needlessly large.

Under a stable data law, labeled calibration examples reveal the score distributions used to
choose the sets. Deployment divides the evidence between environments: source labels describe
prior model behavior, while target inputs show which feature values now occur. Neither reveals
how target labels are distributed within those target features. The missing object is the target
score law within each true class, whose recovery depends on the assumptions connecting the
domains.

Deployment shift makes this information problem concrete. A separate source threshold for each
class gives classwise coverage when source and target scores remain exchangeable within that
class. Once this relation breaks, even unlimited unlabeled target inputs reveal only the target
feature law; they do not determine the score distribution conditional on the unseen true label.
The practical question is then also a learning-theoretic one:

\smallskip
\noindent\textbf{Central question.}
When can labeled source data and unlabeled target inputs determine useful class-conditional
prediction sets, and what target-label budget is needed when they cannot?
\smallskip

Coverage alone does not settle this question. A rule that always returns every label covers every
class without observing a target label. The difficulty is to retain useful sets. We express
usefulness through recovery of the target class quantiles that define the smallest classwise
thresholds, and we also show how threshold error controls excess expected set size under a mild
sensitivity condition. Figure~\ref{fig:alt-overview} summarizes the two information regimes.

\begin{figure}[t]
\floatconts
{fig:alt-overview}
{\caption{Two information regimes for class-conditional prediction under shift. Source labels
and unlabeled target inputs do not in general identify the target classwise score laws; a labeled
target audit estimates their relevant quantiles and supports finite, accurate thresholds. NTU-60
illustrates the source-only failure.}}
{\includegraphics[width=.98\linewidth,trim=0 4bp 0 0,clip]{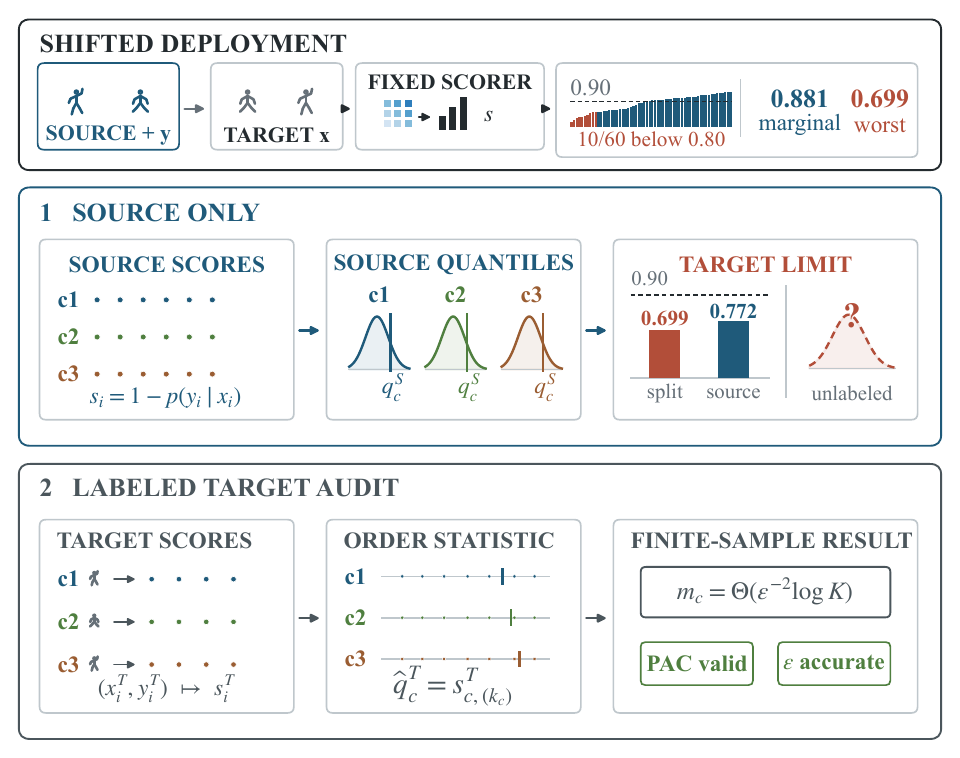}}
\end{figure}

Our analysis has two parts. First, we compare target laws that have the same labeled source law
and the same unlabeled target feature law. Their hidden label mechanisms can induce different
classwise score quantiles. A label-free rule cannot distinguish these laws; covering both forces a
strict set-size cost on one of them. This result states exactly what is missing from the observed
information and why more unlabeled examples do not repair it.

Second, we add a labeled target audit. A binomial-tail order statistic gives simultaneous
high-confidence classwise coverage for any audit size, using an infinite threshold when the data
do not support a finite one. Under local regularity, finite and accurate thresholds require
$O(\varepsilon^{-2}\log(K/\delta))$ labels per class. For fixed, nonadaptive class-stratified
counts of complete labeled pairs, a product-testing construction gives the matching total
$\Omega(K\varepsilon^{-2}\log K)$ dependence at fixed confidence, and the per-class rate under
equal allocation. If labels instead arrive from an
unstratified target stream, the draw budget also depends on class prevalence. We prove that its
$K\varepsilon^{-2}\log K$ order is necessary when the foreground class probabilities are all of
order $1/K$. Thus the logarithmic class dependence is not only a union-bound artifact, and sample
acquisition has a separate class-prevalence cost.

Experiments serve two roles. Controlled studies test the predicted scaling and the effect of
random class counts. Real shifts show the phenomenon that motivates the theory: good marginal
coverage can coexist with poor lower-tail class coverage, and source classwise calibration can
either help or fail depending on the shift. The experiments do not turn the worst-case result
into a claim about every deployment; they show that the identified failure mode occurs in common
models and data shifts.

Our contributions are:
\begin{itemize}[leftmargin=1.35em,itemsep=2pt,topsep=3pt]
  \item Under one deployment observation model, we connect label-free nonidentifiability, its
  set-size cost, and the target labels needed to recover all classwise score quantiles.
  \item We prove that labeled source data and unlimited unlabeled target inputs cannot in general
  identify efficient classwise thresholds under unrestricted joint shift. The result covers arbitrary
  measurable, possibly randomized label-free prediction-set rules.
  \item We characterize the fixed-stratified audit cost for complete labeled target pairs: at fixed confidence, the total complexity
  is $\Theta(K\varepsilon^{-2}\log K)$, and equal allocation requires
  $\Theta(\varepsilon^{-2}\log K)$ labels per class. For a mixed stream, we give a general upper
  bound and a matching lower bound when foreground class probabilities are of order $1/K$.
  \item We connect the theory to controlled and real distribution shifts, showing when source
  calibration helps and when target labels become necessary.
\end{itemize}

%% file: sections/alt_related.tex
\section{Related work}
\label{sec:related}

\paragraph{Conformal prediction under shift.}
Split conformal prediction gives marginal coverage when calibration and test examples are
exchangeable \citep{vovk2005,lei2018}. Weighted conformal prediction extends this guarantee to
covariate shift when the source-to-target density ratio is known
\citep{tibshirani2019}; other methods treat label shift \citep{podkopaev2021labelshift}, quantify
departures from exchangeability \citep{barber2021limits,barber2023}, or protect against a stated
neighborhood of laws \citep{cauchois2024robust}. PAC methods have also been developed for
covariate and label shift \citep{park2022covariate,si2024labelshift}. Recent work extends these
positive results to structured posterior drift \citep{wang2025posteriordrift} and high-dimensional
covariate shift without direct density-ratio estimation \citep{joshi2025likelihood}. More general
nonexchangeable guarantees can also be derived when the relevant joint law is specified
\citep{prinster2024anydistribution}. These results use a known or estimable relation between the
source and target laws. Our lower bound concerns unrestricted joint shift and identifies the target
labels needed when that relation is absent.

\paragraph{Classwise and high-confidence coverage.}
Mondrian conformal prediction calibrates groups separately and gives classwise coverage under
within-class exchangeability \citep{vovk2003mondrian,lofstrom2015bias}. Grouping related classes
can improve efficiency when per-class samples are small \citep{ding2023classconditional}.
Rank-based calibration can further reduce set size for many or imbalanced classes
\citep{shi2024rankcalibration}, while Kandinsky conformal prediction supports overlapping and
fractional groups defined by labels and covariates \citep{bairaktari2025kandinsky}.
Training-conditional and PAC prediction sets provide guarantees that hold with high probability
over calibration data \citep{vovk2012conditional,park2020pac,bian2023trainingconditional}. Our
audit uses a classical binomial tolerance limit \citep{wilks1941,clopper1934}. The new point is the
separation between label-free identification and labeled recovery, together with a matching
many-class lower bound in this deployment setting.

\paragraph{Target audits under unknown shift.}
Audited conformal prediction also uses labeled target data under unknown shift
\citep{zhou2026audited}. It trains an auxiliary model to predict when a fixed classifier will fail,
then uses that signal to improve marginal coverage behavior or to calibrate audit-defined groups.
We study a different target: the true-label score quantile of every class. Our audit estimates
these quantiles directly, without learning a failure predictor, and our main results give an
identification boundary and matching fixed-stratified label-complexity bounds for complete labeled pairs, with explicit
dependence on the number of classes.

\paragraph{Adaptation with unlabeled targets.}
Test-time conformal methods can adapt scores or thresholds using unlabeled target inputs
\citep{yilmaz2022testtime,kasa2025adapting,angelman2025sfcp}. In particular, uncertainty-based
score adaptation can improve prediction sets on shifts with useful structure
\citep{kasa2025adapting}; a recent preprint gives coverage bounds for pseudo-calibration under a
bounded label-conditional covariate shift model \citep{siahkali2026pseudocal}. Our result does not
say that such methods fail on every shift. It shows that no label-free rule can recover all
true-class quantiles uniformly over unrestricted joint shift. The construction keeps every
label-free observation fixed while changing these quantiles, thereby isolating the information
that target labels supply. This view is consistent with domain-adaptation lower bounds
\citep{bendavid2010impossibility}. It is also complementary to testable learning under shift,
which obtains positive guarantees when source and target samples pass a problem-specific test
\citep{klivans2024testable}.

The matrix below summarizes the closest distinctions. The comparison separates the
observation model from the coverage goal and from the type of theoretical conclusion.

\begin{center}
\begin{minipage}{.98\linewidth}
\centering
\small\textbf{Related-work matrix.}
Scope of the closest classwise and target-audit work. A dash means that the paper does not pursue
a lower bound for the deployment question studied here.
\par\vspace{4pt}
\footnotesize
\setlength{\tabcolsep}{2.3pt}
\renewcommand{\arraystretch}{1.03}
\begin{tabular*}{\linewidth}{@{\extracolsep{\fill}}L{.14\linewidth}L{.20\linewidth}L{.17\linewidth}L{.25\linewidth}L{.17\linewidth}@{}}
\toprule
Work & Observed data & Target labels & Coverage and efficiency & Lower-bound scope \\
\midrule
\citet{ding2023classconditional} & I.i.d. calibration and test laws & Labeled calibration sample & Class or cluster coverage; empirical set size & --- \\
\citet{shi2024rankcalibration} & I.i.d. calibration and test laws & Labeled calibration sample & Classwise coverage; rank-based set size & --- \\
\citet{bairaktari2025kandinsky} & I.i.d. calibration and test laws & Labeled calibration sample & Overlapping label and covariate groups & Conditional-coverage rate \\
\citet{zhou2026audited} & Fixed legacy model; unknown target shift & Split target audit & Marginal or audit-group coverage; mean set size & --- \\
This work & Source joint law and target inputs & None, or a classwise audit & Every true class; quantile and excess-size control & Label-free cost; stratified and balanced-stream $K$ rates \\
\bottomrule
\end{tabular*}
\end{minipage}
\end{center}

The matrix separates the target information observed, the required coverage event, and the scope
of the conclusion; these differences organize the comparison.

%% file: sections/alt_problem.tex
\section{Problem and information model}
\label{sec:problem}

Let $\mathcal Y=\{1,\ldots,K\}$ and let $s(x,c)\in\mathbb R$ be a fixed nonconformity score for
assigning class $c$ to input $x$. Lower scores indicate better agreement. A vector of thresholds
$q=(q_1,\ldots,q_K)\in(\mathbb R\cup\{+\infty\})^K$ defines
\[
  C_q(x)=\{c:s(x,c)\le q_c\}.
\]
The source joint law $P_S(X,Y)$ is observed through labeled data. The target feature law
$P_T(X)$ is available through an arbitrarily large unlabeled sample. Unless stated otherwise,
the target conditional law $P_T(Y\mid X)$ is unrestricted. The score is held fixed so that the
analysis concerns calibration, not representation learning.

For class $c$, define the target true-class score distribution and its smallest valid threshold
by
\[
F_c^T(t)=P_T\{s(X,c)\le t\mid Y=c\},\qquad
q_c^\star=\inf\{t:F_c^T(t)\ge1-\alpha\}.
\]
For continuous class-score laws, $q^\star$ gives the smallest valid deterministic rule in the
fixed-score threshold family. With $G_c(t)=P_T\{s(X,c)\le t\}$, define its expected-size benchmark by
\[
  \mathrm{OPT}_s(P_T)
  :=\E_{P_T}|C_{q^\star}(X)|
  =\sum_{c=1}^K G_c(q_c^\star).
\]
The desired classwise property is
\[
P_T\{Y\in C_q(X)\mid Y=c\}\ge1-\alpha
\quad\text{for every }c.
\]
Marginal coverage is the class-frequency-weighted average of these terms and therefore does not
imply this condition.

\paragraph{Validity and usefulness.}
A threshold of $+\infty$ is always valid for a class, so validity alone has zero target-label
complexity. We call a threshold vector $\varepsilon$-accurate when
\[
\mathcal E_\varepsilon(P_T)
=\left\{\max_c|\hat q_c-q_c^\star|\le\varepsilon\right\},
\]
and write
\[
\mathcal V(P_T)
=\left\{F_c^T(\hat q_c)\ge1-\alpha\ \text{for every }c\right\}
\]
for simultaneous validity. Our scale-free usefulness criterion is the probability excess
$F_c^T(\hat q_c)-(1-\alpha)$. We write
\[
\mathcal P_\eta(P_T)=
\left\{0\le F_c^T(\hat q_c)-(1-\alpha)\le\eta\ \text{for every }c\right\}.
\]
This event is unchanged by strictly increasing transformations of the score. When the classwise
score density is bounded below near $q_c^\star$, probability control also gives the threshold
recovery event above. Under a local Lipschitz condition, threshold recovery in turn bounds the
extra expected set size (Corollary~\ref{cor:labelcomplexity_size_main}). Thus the implication chain
is probability-scale recovery, threshold recovery under a density floor, and expected-size control
under marginal Lipschitz continuity.

\paragraph{Three information regimes.}
We compare: (i) source calibration, using labeled source examples only; (ii) label-free target
adaptation, using the source law and $P_T(X)$ but no target labels; and (iii) a target audit, which
adds independent labeled target examples. The first two regimes may use unlimited data. This
choice makes the lower bound an identification result rather than a finite-source-sample effect.

For reference, source Mondrian calibration has the following familiar guarantee.

\begin{proposition}[Source Mondrian coverage]
\label{prop:mondrian}
Suppose the $n_c$ source calibration scores of class $c$ are exchangeable with a target test score
from that class. Let $\hat q_c$ be their $\lceil(n_c+1)(1-\alpha)\rceil$-th order statistic, or
$+\infty$ if the index exceeds $n_c$. Then
\[
1-\alpha\le P\{c\in C_{\hat q}(X)\mid Y=c\}
\le1-\alpha+\frac{1}{n_c+1},
\]
where the upper bound assumes almost surely distinct scores.
\end{proposition}

After joint shift, within-class exchangeability need not hold. Source Mondrian can still work,
but its target coverage must then be measured or justified by an additional shift assumption.

%% file: sections/alt_identification.tex
\section{What label-free data can identify}
\label{sec:identification}

The unlabeled target sample identifies $P_T(X)$, while a classwise threshold depends on
$P_T(X\mid Y=c)$. Under unrestricted joint shift, the latter is not determined by the former.
The next proposition turns this observation into an efficiency cost.

For a target law $P$, let $\widehat C$ denote a generic measurable, possibly randomized prediction
set, and let $\mathrm{OPT}(P)$ be the infimum expected size of such rules that cover every class at
level $1-\alpha$ under $P$.
This benchmark ranges over all prediction-set rules. It can therefore be smaller than
$\mathrm{OPT}_s(P)$, which is the benchmark for the fixed-score threshold family studied by the
constructive audit. Proposition~\ref{prop:impossible} deliberately uses the broader comparison;
the positive results do not claim to recover the global optimum.

\Needspace{8\baselineskip}
\begin{proposition}[Cost of label-free classwise coverage]
\label{prop:impossible}
Fix $P_S$, $s$, $1-\alpha$, and $K\ge2$. Let $Q$ be a target feature law, and choose a class $c$
and disjoint regions $A,B$ with positive $Q$-mass. If $q_B$ is the $(1-\alpha)$ quantile of
$s(X,c)$ under $Q(\cdot\mid B)$ and, for some $\gamma>0$,
\[
  \inf_{x\in A}s(x,c)\ge q_B+\gamma,
\]
then there are two target laws $P_T,P_T'$ consistent with the same observed pair $(P_S,Q)$ such
that
\[
q_c^\star(P_T)=q_B,
\qquad
q_c^\star(P_T')\ge q_B+\gamma.
\]
Any possibly randomized label-free rule $\widehat C$ that covers every class under both laws
satisfies
\[
\E_{P_T}|\widehat C(X)|\ge\mathrm{OPT}(P_T)+\beta
\]
for a constant $\beta>0$ determined by the construction.
\end{proposition}

\paragraph{Proof idea.}
Both laws use $Q$ as their target feature distribution. On region $B$, they assign class $c$ in
the same way. On region $A$, only $P_T'$ assigns class $c$. The true-class scores on $A$ are
larger than the relevant quantile on $B$, so $P_T'$ needs a larger threshold. A label-free rule
has the same output distribution under both laws. If it includes class $c$ often enough on $A$
to cover $P_T'$, those inclusions are unnecessary under $P_T$ and increase expected set size.
This conclusion also covers randomized rules: their internal randomness can be coupled so that
the same observed input produces the same output under the two laws. The loss is not caused by
poor estimation of the feature law. It comes from the unseen allocation of labels across $A$ and
$B$, which changes the true-class score law without changing any label-free observation.
Appendix~\ref{supp:impossible_proof} gives the construction and the explicit value of $\beta$.

\paragraph{What the result does and does not say.}
The proposition is an identification statement for unrestricted joint shift. It does not rule
out label-free guarantees under covariate shift, label shift, a bounded joint density ratio, or
another assumption that links $P_S$ and $P_T$. Those assumptions shrink the set of target laws
compatible with the observations. Without such a link, more unlabeled target inputs estimate
$Q$ more precisely but cannot reveal which compatible label mechanism generated the data.
The positive results under structured shifts are therefore consistent with the proposition: they
restrict the compatible target family before any target label is collected. Under unrestricted
shift, the target feature sample can determine $Q$ arbitrarily well while leaving the relevant
class-conditional score law unknown.
Appendix~\ref{app:label_shift} gives a controlled boundary check. Under pure label shift, the
class-conditional score laws are unchanged, and Source Mondrian remains close to nominal coverage
on every matched class.

%% file: sections/alt_audit.tex
\section{Target auditing and label complexity}
\label{sec:audit}

A target audit observes $m_c$ independent true-class scores from each class $c$. The following
procedure uses a one-sided binomial tolerance limit. It returns an infinite threshold when the
audit is too small to support a finite threshold at the requested confidence.

\begin{algorithm}[htbp]
\floatconts
{alg:pac-audit}
{\caption{PAC target audit for classwise thresholds}}
{%
\DontPrintSemicolon
\KwIn{Audit scores $S_c=\{s(X_i,c):Y_i=c\}$, level $\alpha$, confidence $\delta$}
\KwOut{Thresholds $(\hat q_c)_{c=1}^K$}
\begin{minipage}[t]{.47\linewidth}
\raggedright
\textbf{1. Check each class.}\par\smallskip
Set $\delta_c\leftarrow\delta/(2K)$ and sort
$S_{c,(1)}\le\cdots\le S_{c,(m_c)}$. If
$(1-\alpha)^{m_c}>\delta_c$, return
$\hat q_c\leftarrow+\infty$.
\end{minipage}\hfill
\begin{minipage}[t]{.48\linewidth}
\raggedright
\textbf{2. Select a finite threshold.}\par\smallskip
Otherwise set
$k_c\leftarrow\min\{k\in\{1,\ldots,m_c\}:\Pr[\mathrm{Bin}(m_c,1-\alpha)\ge k]\le\delta_c\}$
and return $\hat q_c\leftarrow S_{c,(k_c)}$.
Repeat for $c=1,\ldots,K$.
\end{minipage}
\par\smallskip
\textbf{Finite-threshold boundary:}\quad
$m_c\ge m_{\mathrm{finite}}:=
\left\lceil\log\{\delta/(2K)\}/\log(1-\alpha)\right\rceil$.
}
\end{algorithm}
\FloatBarrier

Below the stated boundary, the infinite threshold keeps the classwise guarantee valid but makes that
class uninformative. Thus validity is available for every audit size, while a useful finite rule
requires enough labels from each class.

\begin{mdframed}[style=altkeyresult]
\begin{theorem}[Target-label cost of useful classwise thresholds]
\label{thm:labelcomplexity}
Fix $\alpha,\delta\in(0,1)$. Parts (a)--(b) hold for any target law with all classes present;
parts (c)--(d) are worst-case lower bounds. Probabilities include the audit and any internal
randomization.
\begin{enumerate}[label=(\alph*),leftmargin=1.5em,itemsep=2pt,topsep=3pt]
  \item For any class counts, including zero, Algorithm~\ref{alg:pac-audit} satisfies
  $P\{\mathcal V(P_T)\}\ge1-\delta$.
  \item If every $F_c^T$ is continuous and has density bounded below on a right neighborhood of
  $q_c^\star$, then fixed class-stratified counts of
  $O(\varepsilon^{-2}\log(K/\delta))$ scores per class suffice for
  $P\{\mathcal V(P_T)\cap\mathcal E_\varepsilon(P_T)\}\ge1-\delta$.
  \item For sufficiently small $\varepsilon$, there are two target laws sharing
  $(P_S,P_T(X))$ such that any classwise-threshold procedure with at most
  $c_0\varepsilon^{-2}$ independent labeled target pairs fails
  validity or accuracy with probability at least $1/4$ on one law.
  \item For sufficiently small $\varepsilon$ and $K\ge4$, there is an instance with
  $\lfloor K/2\rfloor$ informative class pairs such that any classwise-threshold procedure using
  fixed, nonadaptive class-stratified labeled pairs and having uniform success probability $3/4$
  requires
  \[
    \sum_{c=1}^{K}m_c=\Omega(K\varepsilon^{-2}\log K).
  \]
  Under equal class counts, every class count has order
  $\Omega(\varepsilon^{-2}\log K)$.
\end{enumerate}
\end{theorem}
\end{mdframed}

The order notation in part~(b) is for $\varepsilon\to0$ with $\alpha$ fixed and with positive
lower bounds on the local score densities and neighborhood widths held uniformly across classes.

The two factors in the rate have different roles. The factor $K$ counts the classwise score laws
that must be recovered, while $\log K$ is the cost of making every recovery succeed in one joint
event. Thus the theorem controls the full threshold vector at the stated confidence, not a
collection of unrelated per-class guarantees.

\Needspace{10\baselineskip}
\begin{corollary}[Scale-free probability recovery]
\label{cor:probscale-main}
Assume that every $F_c^T$ is continuous. Fixed class-stratified counts of
$O(\eta^{-2}\log(K/\delta))$ scores per class suffice to ensure, with probability at least
$1-\delta$, that $\mathcal P_\eta(P_T)$ holds; equivalently, simultaneously for every class
\[
  0\le F_c^T(\hat q_c)-(1-\alpha)\le\eta.
\]
This event and the resulting prediction sets are unchanged by any strictly increasing
transformation of the score. The lower bounds in Theorem~\ref{thm:labelcomplexity}(c)--(d) also
hold for this probability-scale criterion after replacing $\varepsilon$ by $\eta$.
\end{corollary}

\paragraph{Why the bounds match.}
The binomial order statistic first controls probability excess above $1-\alpha$, and a union bound
makes the control simultaneous over $K$ classes. This gives
Corollary~\ref{cor:probscale-main} without a density assumption. A local density floor then converts
probability error into the threshold error in Theorem~\ref{thm:labelcomplexity}(b). For the lower
bound, pair the classes and attach one hidden sign $a_j\in\{-\tau,+\tau\}$ to each pair through
the score CDF $F_a(t)=t+a(t-t^2)$, with $\tau$ proportional to $\varepsilon$. Flipping the sign
swaps the two class-conditional score laws in that pair, and the relevant quantiles are more than
$2\varepsilon$ apart. Simultaneous validity and accuracy therefore decode every sign even when the
procedure sees complete labeled pairs. If $n_j$ is the combined sample count in pair $j$, its
testing error is bounded below by a constant times $\exp(-C\varepsilon^2n_j)$. Requiring the whole
sign vector to be correct with constant probability forces
$\sum_j\exp(-C\varepsilon^2n_j)=O(1)$, and hence total cost
$\Omega(K\varepsilon^{-2}\log K)$. In a mixed stream, each foreground coordinate appears with
probability of order $1/K$, which yields the same lower-bound order for the number of labeled draws.
Appendix~\ref{supp:labelcomplexity_proof} gives the complete argument and constants.

\begin{corollary}[Labels drawn from an unstratified target stream]
\label{cor:stream-acquisition}
Let $\pi_c=P_T(Y=c)>0$, and let $b_c$ be deterministic class counts that make the audit succeed
with probability at least $1-\delta/2$. If $N$ target examples are labeled independently, then
the same conclusion holds with probability at least $1-\delta$ whenever
\[
  N\ge \max_c\frac{\max\{2b_c,\ 8\log(2K/\delta)\}}{\pi_c}.
\]
\end{corollary}

\begin{corollary}[Mixed-stream lower bound]
\label{cor:stream-lower}
For sufficiently small $\varepsilon$ and $K\ge4$, there is an instance with $K-1$ foreground
classes, each with probability $1/\{2(K-1)\}$, such that any classwise-threshold procedure using
$N$ independent labeled target pairs and succeeding on simultaneous validity and threshold
recovery with probability at least $3/4$ for every law in the instance family requires
\[
  N=\Omega(K\varepsilon^{-2}\log K).
\]
The procedure may also use the full source law and unlimited unlabeled target inputs.
\end{corollary}

The two corollaries separate quantile estimation from sample acquisition. The values $b_c$
control estimation after the classwise samples are available, while $1/\pi_c$ controls how many
mixed-stream labels are needed to obtain them. At fixed confidence, the lower bound matches the
upper bound when the foreground class probabilities are all of order $1/K$. It does not assert
tightness for every imbalanced class distribution.

Together, these results provide an information-to-action characterization. They identify the target
quantities that source labels and unlabeled target inputs do not determine, quantify the efficiency
cost of uniform label-free validity, give the fixed-stratified audit cost for recovering the missing
quantiles, and characterize mixed-stream acquisition in the balanced-foreground regime.

\begin{corollary}[Expected set size from threshold recovery]
\label{cor:labelcomplexity_size_main}
Under Theorem~\ref{thm:labelcomplexity}(b), suppose
$G_c(t)=P_T\{s(X,c)\le t\}$ is $L_c$-Lipschitz on
$[q_c^\star,q_c^\star+\varepsilon]$. On the same event,
\[
0\le \E|C_{\hat q}(X)|-\mathrm{OPT}_s(P_T)
\le\sum_{c=1}^K L_c|\hat q_c-q_c^\star|.
\]
\end{corollary}

This corollary is one-sided: it turns accurate thresholds into a bound on excess size. The lower
bound itself concerns classwise threshold recovery, not total expected set size.

%% file: sections/alt_experiments.tex
\section{Experiments}
\label{sec:experiments}

The experiments test the assumptions and consequences of the theory. Within each study, all
compared methods share the same fixed scorer and differ only in the calibration information they
use. The target level is $0.90$ unless stated otherwise. We report marginal coverage,
worst-class coverage, $\cvar$ (the mean over the weakest
tenth of classes), and mean prediction-set size. Full protocols and additional results are in
Appendix~\ref{app:experiments}.

\subsection{Controlled tests of the theory}

The controlled studies isolate the three claims that are difficult to read from real shifts.
Source training, calibration, and evaluation examples are disjoint, and every comparison keeps
the scorer fixed.

\begin{center}
\begin{minipage}{.94\linewidth}
\small
\setlength{\tabcolsep}{4pt}
\renewcommand{\arraystretch}{1.08}
\begin{tabular*}{\linewidth}{@{\extracolsep{\fill}}L{.22\linewidth}L{.27\linewidth}L{.41\linewidth}@{}}
\toprule
Study & Controlled change & Primary check \\
\midrule
Coupled identification & indistinguishable target laws & label-free output and size/coverage cost \\
Scaling & $K,\ \varepsilon,\ \delta$ & transition in $m_c/\{\varepsilon^{-2}\log(K/\delta)\}$ \\
Random counts & balanced to $10{:}1$ priors & finite thresholds under mixed-stream sampling \\
\bottomrule
\end{tabular*}
\end{minipage}
\end{center}

In the coupled study, all 180 label-free output pairs are identical, while the calculated
coverage/size cost agrees with its analytic value within $0.0058$. In the scaling study, 71 of 72
common 90\% transition diagnostics are resolved; the normalized crossings range from $0.837$ to
$2.221$, so a pre-set factor-two band is narrowly missed. In the random-count study with $K=345$, the deterministic
quota is $m_\star=91$ per class. The probability that every threshold is finite reaches $0.90$ at
$N\pi_{\min}/m_\star=1.5$ for all four prior profiles, while at $N=Km_\star$ a $10{:}1$ prior leaves
$0.590$ of thresholds infinite on average. These fixed-run outcomes support the predicted scale
and also show that its constants and the cost of acquiring rare classes matter.

\subsection{Evidence under real shifts}

We now move from controlled checks of the theory to diagnostic evidence under real distribution
shifts.

Table~\ref{tab:alt-real-shifts} compares a shared source threshold (Split) with separate source
thresholds (Mondrian). The action-recognition studies use SPD sequence descriptors with ST-GCN
features; the image studies use fixed CLIP classifiers. For this table, class coverage is first
averaged over ten source/calibration seeds and then summarized across classes.

\begin{center}
\begin{minipage}{.94\linewidth}
\centering
\refstepcounter{table}\label{tab:alt-real-shifts}%
\small\textbf{Table~\thetable:}
Aggregate class-coverage profiles under representative shifts. Entries report Split CP $\to$ Source
Mondrian for marginal coverage, $\cvar$, worst-class coverage, and mean set size. C, R, and S
denote Clipart, Real, and Sketch. Classwise coverages are averaged over the ten seeds before the
$\cvar$ and worst-class functionals are applied.
\par\vspace{5pt}
\small
\setlength{\tabcolsep}{2.6pt}
\renewcommand{\arraystretch}{1.00}
\begin{tabular*}{\linewidth}{@{\extracolsep{\fill}}L{0.27\linewidth}rcccc@{}}
\toprule
Setting & $K$ & Marginal & $\cvar$ & Worst & Size \\
\midrule
NTU subject & 60 & $0.881\to0.881$ & $0.735\to0.810$ & $0.699\to0.772$ & $3.85\to4.74$ \\
NTU view & 60 & $0.924\to0.928$ & $0.826\to0.897$ & $0.803\to0.875$ & $4.17\to5.21$ \\
NTU $\to$ N-UCLA & 5 & $0.501\to0.494$ & $0.000\to0.001$ & $0.000\to0.001$ & $0.72\to0.75$ \\
CIFAR-100 corruption & 100 & $0.838\to0.824$ & $0.564\to0.706$ & $0.387\to0.669$ & $2.56\to3.61$ \\
Office-Home C $\to$ R & 65 & $0.981\to0.974$ & $0.815\to0.893$ & $0.192\to0.864$ & $1.56\to3.31$ \\
DomainNet S $\to$ R & 345 & $0.972\to0.945$ & $0.869\to0.780$ & $0.456\to0.617$ & $4.88\to13.93$ \\
\bottomrule
\end{tabular*}
\end{minipage}
\end{center}

\Needspace{8\baselineskip}
The table shows three distinct outcomes. On NTU subject/view and CIFAR-100 corruption, source
Mondrian raises lower-tail coverage with a moderate increase in set size. On NTU to N-UCLA, it
does almost nothing. On DomainNet, the weakest class improves while the broader lower tail falls
and the sets grow sharply. No statistic available from the source thresholds alone determines in
advance which outcome will occur; this is consistent with the identification limit, though it is
not itself a proof of that limit.

\begin{figure}[t]
\centering
\begin{minipage}{.94\linewidth}
\floatconts
{fig:alt-budget}
{\caption{NTU-60 cross-subject target-label study with one fixed source split and 200 paired target
audit/evaluation resamples: (a) lower-tail class coverage and (b) mean set size.}}
{\includegraphics[width=\linewidth]{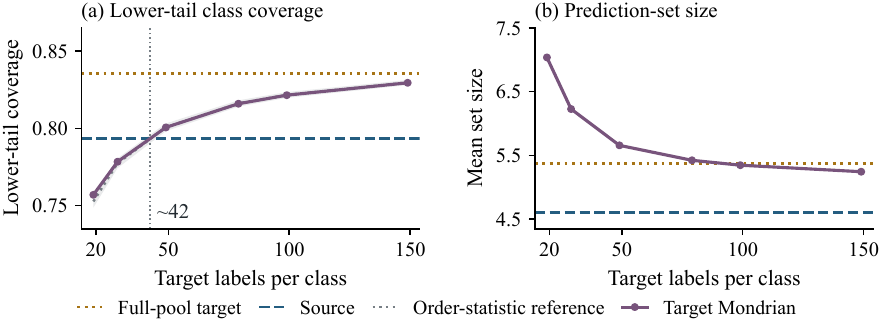}}
\end{minipage}
\end{figure}

Figure~\ref{fig:alt-budget} uses source seed and split 0, then repeats paired target audit and
held-out evaluation resampling 200 times. Within each resample, $\cvar$ sorts the 60 held-out class
coverages and averages the lowest six; the curve then averages this value across resamples. This
protocol is separate from the ten-seed profile in Table~\ref{tab:alt-real-shifts}; its source
reference is therefore $0.793$ with mean size $4.61$. The full-pool target reference is $0.836$,
not $0.90$, because it applies the same lower-tail functional after finite held-out evaluation and
is not a population coverage guarantee.
At $m=19$, ordinary Target Mondrian has $\cvar=0.757$ and mean size $7.04$, and it passes the
source reference near $m=42$. PAC auditing asks for a stronger simultaneous statement. With
$K=60$, $\alpha=.10$, and $\delta=.05$, all thresholds first become finite at $m=74$ per class.
At the evaluated finite budgets $m=79,99,149$, its sets are respectively $4.425$, $2.863$, and
$2.250$ times as large as ordinary Target Mondrian. The cost falls with more labels, but it remains
visible because the audit protects every class with high confidence, not only average empirical
performance.

Figure~\ref{fig:alt-empirical-summary} shows how the aggregate results arise. Panel (a) sorts each
method's 60 NTU cross-subject class coverages independently, so the curves compare coverage
distributions rather than matched classes. Weighted CP shifts much of the distribution upward and
Source Mondrian raises its lower end, but neither source-only rule keeps every class near the
target. Panel (b) compares view, subject, and NTU-to-N-UCLA sensor shifts; marginal coverage falls,
while the weakest class falls faster and nearly reaches zero. Panels (c)--(d)
show that changing the target level or the LAC/APS/RAPS score does not remove the source-only
shortfall. The labeled target oracle in panel (a) is a reference, not a deployable source-only
method.

These comparisons separate three levels of evidence. Class profiles reveal which classes fail,
but they do not by themselves provide a finite-sample guarantee. Ordinary Target Mondrian uses
labeled target scores and can give smaller sets, but its empirical quantiles do not provide the
joint confidence statement studied here. The PAC audit adds that statement through conservative
order statistics; with a small audit, the price is an infinite threshold or a larger set. Source
methods occupy a different regime: they require no target labels, but their success depends on a
relation between the source and target classwise score laws. The figures therefore compare
different information regimes, not simply different implementations of the same guarantee.

\begin{figure}[t]
\centering
\begin{minipage}{.94\linewidth}
\floatconts
{fig:alt-empirical-summary}
{\caption{Distribution and robustness of classwise failure. (a) NTU-60 cross-subject class
profiles. (b) Source Mondrian across view, subject, and cross-dataset sensor shifts. (c) Coverage targets with
ST-GCN. (d) LAC, APS, and RAPS scores with CTR-GCN.}}
{\includegraphics[width=\linewidth]{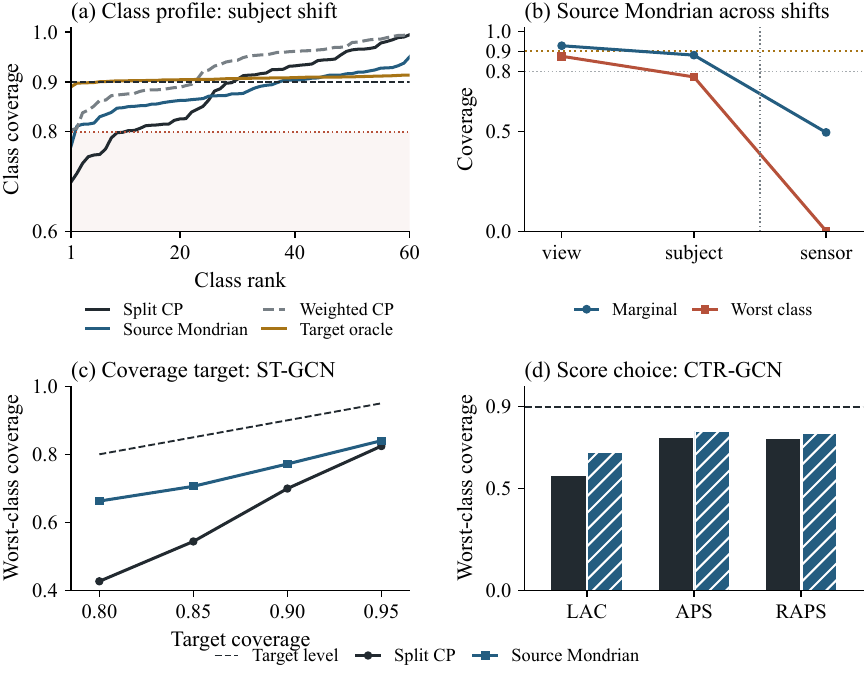}}
\end{minipage}
\end{figure}

%% file: sections/alt_discussion.tex
\section{Discussion and limitations}
\label{sec:discussion}

The theory and experiments draw the same boundary. Marginal success need not imply classwise
success, and source calibration is useful only when the shift relation supports it. A deployment
should therefore state that relation first. If a verified covariate-, label-, or other structured
shift identifies the target score laws, a label-free method may be justified. Under unrestricted
joint shift, unlabeled inputs describe feature shift but cannot certify the missing classwise
quantiles; a labeled target audit supplies that information. Its finite-threshold boundary marks
when an informative rule becomes possible, and its recovery bound marks the larger budget needed
for accurate thresholds.

Several limits are important. The many-class lower bound assumes fixed, nonadaptive stratified
counts and fixed confidence. For a mixed stream, Corollaries~\ref{cor:stream-acquisition} and
\ref{cor:stream-lower} match in order when foreground probabilities are of order $1/K$; we do not
claim tightness for every imbalance or adaptive acquisition rule. Threshold error depends on score
scale, although the probability-scale result is invariant to monotone transformations, and the
lower bound does not match total expected set size. The experiments span several action and image
shifts but not every target law or operational cost.

\section{Conclusion}

Useful class-conditional sets require information about the target true-class score laws. Under
unrestricted joint shift, source labels and unlabeled target inputs do not identify them; a labeled
audit does. At fixed confidence and under local regularity, a fixed stratified audit costs
$\Theta(K\varepsilon^{-2}\log K)$ labels in total, or
$\Theta(\varepsilon^{-2}\log K)$ per class under equal allocation. A mixed stream adds a separate
class-prevalence acquisition cost. The resulting workflow is direct: state the shift assumption,
check whether the observed data identify the target classwise score laws, and collect target labels
when they do not.

%% file: sections/alt_proofs.tex
\section{Proofs and extensions}
\label{app:proofs}

\subsection{Proof of the impossibility proposition}
\label{supp:impossible_proof}

The main paper states the impossibility proposition for label-free classwise coverage under joint
shift; its construction and proof follow.

\medskip
\noindent\emph{Construction.} Let $Q$ be the feature law in the main-paper proposition,
and put $m=Q(A)>0$. On $B$, assign class $c$ with a constant probability
$\mu_c/Q(B)$ under both target laws, where $0<\mu_c<Q(B)$ is small enough that
\[
\theta:=\frac{m}{m+\mu_c}>\alpha.
\]
Conditional on $Y=c$ and $X\in B$, the score therefore has the distribution of
$s(X,c)$ under $Q(\cdot\mid B)$, whose $(1-\alpha)$ quantile is $q_B$.
$P_T'$ assigns all of the $A$-mass to $Y{=}c$, while $P_T$ assigns none of it to $c$;
outside $A\cup B$, neither law assigns $c$. Remaining mass can be distributed among
the other classes so that each has positive probability. Thus,
\begin{align*}
q_c^\star(P_T')&\ge q_B+\gamma, \\
q_c^\star(P_T)&=q_B.
\end{align*}
Indeed, under $P_T'$ the conditional mass of $A$ given $Y=c$ is $\theta>\alpha$. Below
$\inf_{x\in A}s(x,c)$, the class-$c$ score CDF is therefore at most $1-\theta<1-\alpha$, which
proves the first inequality; under $P_T$, the class-$c$ score law is exactly the law on $B$.
Both joints share $Q$ and are consistent with $P_S$, since label-free observables place no
restriction on $P_T(Y\mid X)$; class $c$ has positive mass under both. Validity under $P_T'$ forces
\[
\Pr[c\in\hat C(X)\mid X\in A]\ge\frac{\theta-\alpha}{\theta}.
\]
Under $P_T$ these inclusions are spurious. The rule is assumed per class valid on $P_T$, and
deleting them preserves that validity (class $c$ has no $A$-mass there), so the deleted rule is admissible in the infimum defining $\mathrm{OPT}(P_T)$, and the deletion cuts expected size by at least
\[
\beta=\frac{m(\theta-\alpha)}{\theta}.
\]
$\qed$

\medskip
\noindent\emph{Relation to classical impossibility results.} The classical impossibility of distribution free \emph{conditional} coverage \citep{vovk2012conditional,leiwasserman2014,barber2021limits} stems from the nonatomicity of $X$ within a single unknown i.i.d.\ law (attainable by Mondrian CP given labels). Our obstruction is an identification failure \emph{across} laws, in that $(P_S,P_T(X))$ does not identify the target class conditional score law under unrestricted
$P_T(Y\mid X)$, at the same set-size inflation price that worst-case robust conformal methods pay
by design \citep{cauchois2024robust}. Group-conditional methods can cover richer families of groups
when labeled calibration data come from the deployment law \citep{bairaktari2025kandinsky}. Their
shift extension imposes structure on the joint source-to-target density ratio. Our identification
question instead starts with labeled source data and unlabeled target inputs, without such a link.

\begin{remark}[Identifying assumptions that change the problem]
\label{cor:impossible}
Valid and efficient label-free target coverage for every class requires an identifying
restriction on $P_T(Y\mid X)$. The main paper's impossibility result shows that none is available
over the unrestricted consistent set; it does not establish that the two regimes below are the only
ones. Examples include covariate shift ($p(y\mid x)$ invariant, so class-conditional weighted CP with
the covariate ratio is valid given $P_T(X)\ll P_S(X)$, and approximate with estimated weights,
\citealp{tibshirani2019}), or label shift ($p(x\mid y)$ invariant with $\pi_c^T>0$, so source
Mondrian applies, \citealp{podkopaev2021labelshift}). Our construction illustrates what can occur
when neither restriction is imposed. Within the constructed compatible family, a rule that is
valid over the unrestricted identified set must protect against an extremal law and may therefore
overcover on another law. A procedure that matches the target oracle on one compatible law cannot
do so uniformly over the family. The cross-sensor NTU$\to$N-UCLA experiment is a plausible example
of joint shift, but its threshold failures do not identify the shift mechanism. Stronger guarantees
require additional assumptions whose slack is not determined by $(P_S,P_T(X))$ alone.
\end{remark}

\subsection{Proof of the label complexity theorem}
\label{supp:labelcomplexity_proof}

\paragraph{Scope of the rate.} The $\varepsilon^{-2}\log K$ rate is familiar for estimating
$K$ quantiles at once: DKW and a union bound give the upper bound, and product
testing gives a matching lower bound \citep{yu1997lecam,tsybakov2009}. The
class threshold selected by an exact binomial confidence limit is also a
classical one-sided tolerance limit \citep{wilks1941,clopper1934}, with a
standard training-conditional reading in conformal prediction
\citep{vovk2012conditional,park2020pac,bian2023trainingconditional}.

Here the rate is one part of an information-to-action characterization for a
single joint PAC validity event. The analysis identifies which target
class-conditional score laws are missing from the label-free observations,
quantifies the efficiency cost of uniform label-free validity, and gives the
target-label cost of recovering their quantiles. The lower bound allows complete labeled pairs and
matches the total fixed-stratified cost and the per-class cost under equal allocation. A second reduction
matches the mixed-stream upper bound when the foreground class probabilities are of order $1/K$.

We prove Theorem~\ref{thm:labelcomplexity} for the single validity notion stated in the paper. For a
target law $P$, define
\[
\mathcal V(P)=
\left\{F_c^T(\hat q_c)\ge1-\alpha\ \text{for every }c\right\},
\qquad
\mathcal E_\varepsilon(P)=
\left\{\max_c|\hat q_c-q_c^\star|\le\varepsilon\right\}.
\]
Both are events over the labeled audit and any internal randomization. Every upper and lower bound
below concerns $P_{\mathcal A}\{\mathcal V(P)\cap\mathcal E_\varepsilon(P)\}$ for a classwise
threshold predictor.

\subsection*{(a) Simultaneous PAC validity}
Condition on the audit counts $(m_c)_c$. The class-$c$ scores
$S_{c,1},\ldots,S_{c,m_c}$ are i.i.d.\ from $F_c^T$, and the per class error budget is
$\delta_c=\delta/(2K)$. Throughout this proof
$N_c\sim\mathrm{Binomial}(m_c,1-\alpha)$.

Take first a continuous $F_c^T$ and $m_c>0$. The variables $U_{c,i}=F_c^T(S_{c,i})$ are then
i.i.d.\ uniform on $[0,1]$ and $F_c^T(S_{c,(k)})=U_{c,(k)}$, whose law is
$\mathrm{Beta}(k,m_c-k+1)$. The Beta to binomial identity states that for every $x\in[0,1]$
\[
P\{\mathrm{Beta}(k,m_c-k+1)\le x\}=P\{\mathrm{Binomial}(m_c,x)\ge k\},
\]
both sides being the probability that at least $k$ of $m_c$ uniform draws fall below $x$.
Evaluating at $x=1-\alpha$, and using that the Beta law has no atoms,
\begin{equation}
P_{\mathcal A}\left\{F_c^T(S_{c,(k)})<1-\alpha\right\}=P\{N_c\ge k\}.
\label{eq:pac_mondrian_exact}
\end{equation}
The tail $P\{N_c\ge k\}$ is non-increasing in $k$ and equals $(1-\alpha)^{m_c}$ at $k=m_c$, so the
set $\{k\le m_c:P\{N_c\ge k\}\le\delta_c\}$ is nonempty exactly when
$(1-\alpha)^{m_c}\le\delta_c$, which is the finite-threshold condition of the procedure, and $k_c$
is its smallest element. On that event $\hat q_c=S_{c,(k_c)}$, and
\eqref{eq:pac_mondrian_exact} gives
\[
P_{\mathcal A}\left\{F_c^T(\hat q_c)\ge1-\alpha\right\}\ge1-\delta_c,
\]
with equality of the failure probability to the binomial tail. The index $k_c$ is a function of
$m_c$, $\alpha$ and $\delta_c$ alone, never of the scores, which is what makes the fixed order
statistic argument available.

Now drop continuity. Right continuity of $F_c^T$ gives $F_c^T(q_c^\star)\ge1-\alpha$, so
\[
\left\{F_c^T(S_{c,(k_c)})<1-\alpha\right\}\subseteq\left\{S_{c,(k_c)}<q_c^\star\right\}.
\]
The count $N_c^-=\#\{i:S_{c,i}<q_c^\star\}$ is $\mathrm{Binomial}(m_c,p_c^-)$ with
$p_c^-=F_c^T(q_c^{\star-})\le1-\alpha$ by the definition of the infimum, the event
$\{S_{c,(k_c)}<q_c^\star\}$ is exactly $\{N_c^-\ge k_c\}$, and the binomial upper tail is
non-decreasing in the success probability. Hence
\[
P_{\mathcal A}\left\{F_c^T(\hat q_c)<1-\alpha\right\}
\le P\{N_c^-\ge k_c\}
\le P\{N_c\ge k_c\}
\le\delta_c.
\]
Validity therefore survives ties and atoms, and it is strictly conservative whenever
$F_c^T(q_c^{\star-})<1-\alpha$, which is what an atom that crosses the target level produces.

If $(1-\alpha)^{m_c}>\delta_c$, which includes every $m_c=0$ since $\delta_c<1$, PAC Audit
Mondrian sets $\hat q_c=+\infty$, so
$F_c^T(\hat q_c)=1$ and class $c$ contributes no failure probability. A union bound over the $K$
classes therefore gives
\[
P_{\mathcal A}\{\mathcal V(P)\}\ge1-K\delta_c=1-\frac\delta2\ge1-\delta,
\]
without a restriction on $P_T(Y\mid X)$ and with no continuity requirement. The other half of the
error budget is spent on the upper tail in part (b), which is what the factor two in
$\delta_c=\delta/(2K)$ pays for.

For comparison, the ordinary uninflated conformal threshold uses the
$\lceil(m_c+1)(1-\alpha)\rceil$-th audit score, with $+\infty$ when that index exceeds $m_c$.
The index is finite exactly when
\[
\lceil(m_c+1)(1-\alpha)\rceil\le m_c,
\]
which is equivalent to $m_c\ge\lceil1/\alpha\rceil-1$. Below this threshold the $+\infty$
convention remains valid because class $c$ is always included. Thus $m_c=9$ at $\alpha=0.1$ is the
first finite ordinary conformal threshold. It is a non-vacuity threshold and is unrelated to the
existence of a validity guarantee. PAC Audit Mondrian has its own finite-threshold condition
$(1-\alpha)^{m_c}\le\delta_c$. \hfill$\square$

\subsection*{(b) Sufficiency for quantile recovery}
Assume $F_c^T$ is continuous and has density at least $f_{0c}>0$ on
$[q_c^\star,q_c^\star+r_c]$. Fix a class for which $(1-\alpha)^{m_c}\le\delta_c$, so that
$\hat q_c=S_{c,(k_c)}$ is finite. Continuity makes $F_c^T(\hat q_c)=U_{c,(k_c)}$ exactly, with the
$\mathrm{Beta}(k_c,m_c-k_c+1)$ law of part (a). Write
\[
\eta_c=\mathrm{Beta}^{-1}_{k_c,\,m_c-k_c+1}(1-\delta_c)-(1-\alpha)
\]
for the upper tail excess of that law, where $\mathrm{Beta}^{-1}_{a,b}$ is the quantile function of
$\mathrm{Beta}(a,b)$. Since the Beta law has no atoms, the definition of a quantile gives
\begin{equation}
P_{\mathcal A}\left\{F_c^T(\hat q_c)<1-\alpha+\eta_c\right\}\ge1-\delta_c.
\label{eq:pac_mondrian_cdf_upper}
\end{equation}
A union bound over classes bounds the total upper tail failure by $K\delta_c=\delta/2$, and part
(a) spends the other $\delta/2$, so the lower and the upper inequality hold simultaneously for
every class with probability at least $1-\delta$.

Two readings of $\eta_c$ are worth recording, both stated through
\[
e_c=\sqrt{\frac{\log(2K/\delta)}{2m_c}},
\]
which is a convenient Hoeffding margin. A Beta quantile is at most one, so
$\eta_c\le\alpha$ always and the bound $\eta_c\le2e_c$ below is immediate whenever
$2e_c\ge\alpha$. Assume therefore $2e_c<\alpha$. Hoeffding's upper tail gives
$P\{N_c\ge m_c(1-\alpha)+t\}\le e^{-2t^2/m_c}$, and $t=\sqrt{m_c\log(1/\delta_c)/2}=m_ce_c$
satisfies $m_c(1-\alpha)+t<m_c$ under the standing assumption, so this $t$ qualifies in the
definition of $k_c$ and
\[
k_c\le\left\lceil m_c(1-\alpha)+\sqrt{\frac{m_c\log(1/\delta_c)}2}\right\rceil.
\]
In particular $k_c-1\le m_c(1-\alpha)+m_ce_c$.
For continuous $F_c^T$ and any $p$,
$P_{\mathcal A}\{F_c^T(\hat q_c)>p\}=P\{\mathrm{Binomial}(m_c,p)\le k_c-1\}$, and Hoeffding's lower
tail bounds that probability by $\delta_c$ once
\[
p\ge\frac{k_c-1}{m_c}+\sqrt{\frac{\log(1/\delta_c)}{2m_c}}.
\]
The value $p_0=(k_c-1)/m_c+e_c$ satisfies $k_c-1\le m_cp_0$ and, by the previous display,
$p_0\le1-\alpha+2e_c<1$, so Hoeffding applies at $p_0$ and $p_0$ is admissible. The smallest
admissible $p$ is $1-\alpha+\eta_c$ by the definition of the Beta quantile, so $p_0$ yields the
closed form
\begin{equation}
\eta_c\le2\sqrt{\frac{\log(2K/\delta)}{2m_c}}=2e_c.
\label{eq:eta_hoeffding}
\end{equation}
The Beta quantile is the sharper of the two and is what an implementation should use.

Part (a) gives the lower inequality
\[
F_c^T(\hat q_c)\ge1-\alpha=F_c^T(q_c^\star).
\]
The density lower bound makes $F_c^T$ strictly increasing on the stated interval. It follows first
that $\hat q_c\ge q_c^\star$. If
\[
B_c=\frac{\eta_c}{f_{0c}}\le r_c,
\]
then $\hat q_c$ cannot exceed $q_c^\star+B_c$, since otherwise integration of the density lower bound
over $[q_c^\star,q_c^\star+B_c]$ would give
\[
F_c^T(\hat q_c)-F_c^T(q_c^\star)
\ge f_{0c}B_c
=\eta_c,
\]
which contradicts the strict inequality $F_c^T(\hat q_c)<1-\alpha+\eta_c$ that
\eqref{eq:pac_mondrian_cdf_upper} supplies on this event, since subtracting
$F_c^T(q_c^\star)=1-\alpha$ from it gives
$\eta_c\le F_c^T(\hat q_c)-F_c^T(q_c^\star)<\eta_c$. Therefore, simultaneously over classes on the event just described,
\[
1-\alpha\le F_c^T(\hat q_c)\le1-\alpha+\eta_c,
\qquad
0\le\hat q_c-q_c^\star\le\frac{\eta_c}{f_{0c}}\le\frac{2e_c}{f_{0c}}.
\]
This proves the claimed probability $1-\delta$ for validity and quantile recovery under the one
procedure.

Here is one explicit sufficient sample size. Put $h_c=\min\{\varepsilon,r_c\}$ and
$\ell=\log(2K/\delta)$. The two conditions
\[
m_c\ge
\max\left\{
\frac{2\ell}{f_{0c}^2h_c^2},
\frac{\ell}{\log\{1/(1-\alpha)\}}
\right\}
\]
imply $\eta_c\le2e_c\le f_{0c}h_c$ and $(1-\alpha)^{m_c}\le\delta_c$, so the selected threshold is
finite and $0\le\hat q_c-q_c^\star\le h_c\le\varepsilon$. This yields
$m_c=O(\varepsilon^{-2}\log(K/\delta))$ when $\alpha,f_{0c},r_c$ are fixed. The second condition
replaces the looser $\ell/(2\alpha^2)$ condition from a direct DKW construction, and the first
improves its $8\ell/(f_{0c}^2h_c^2)$ constant by a factor of four. The term in
$1/(f_{0c}h_c)$ disappears, since the
exact construction controls the upper tail of $F_c^T(\hat q_c)$ directly. \hfill$\square$

\subsection*{Labels drawn from an unstratified target stream}
Let $N_c$ be the number of examples from class $c$ among $N$ independently labeled target draws.
Then $N_c\sim\mathrm{Binomial}(N,\pi_c)$. If $N\pi_c\ge2b_c$, the event $N_c<b_c$ is contained in
$N_c<N\pi_c/2$. A multiplicative Chernoff bound gives
\[
  P\{N_c<b_c\}\le \exp(-N\pi_c/8).
\]
The condition in Corollary~\ref{cor:stream-acquisition} makes this probability at most
$\delta/(2K)$ for every class. A union bound therefore gives all required class counts with
probability at least $1-\delta/2$. Conditional on those counts, retain the first $b_c$ scores from
each class and run the audit with confidence parameter $\delta/2$. Its success probability is at
least $1-\delta/2$, so another union bound proves the corollary. \hfill$\square$

\subsection*{Expected set size corollary}
\begin{corollary}[Expected set size from threshold recovery]
\label{cor:labelcomplexity_size}
Under part~(b), suppose $G_c(t)=P_T\{s(X,c)\le t\}$ is $L_c$-Lipschitz on
$[q_c^\star,q_c^\star+r_c]$. On the same event of probability at least $1-\delta$,
\[
0\le\E_{P_T}|C_{\hat q}(X)|-\sum_{c=1}^K G_c(q_c^\star)
\le\sum_{c=1}^K L_cB_c,
\]
where $B_c=\eta_c/f_{0c}$ is the threshold-error bound from part~(b).
\end{corollary}

For every deterministic threshold vector $q$,
\[
\mathbb E_{P_T}|C_q(X)|
=\sum_{c=1}^K P_T\{s(X,c)\le q_c\}
=\sum_{c=1}^K G_c(q_c).
\]
On the event proved in part (b), both $q_c^\star$ and $\hat q_c$ lie in the interval on which
$G_c$ is $L_c$-Lipschitz, and $\hat q_c\ge q_c^\star$. Consequently,
\begin{align*}
0
&\le
\mathbb E_{P_T}|C_{\hat q}(X)|
-\sum_{c=1}^K G_c(q_c^\star) \\
&=\sum_{c=1}^K\{G_c(\hat q_c)-G_c(q_c^\star)\} \\
&\le\sum_{c=1}^K L_c(\hat q_c-q_c^\star) \\
&\le
\sum_{c=1}^K\frac{L_c\eta_c}{f_{0c}}
\;\le\;
\sum_{c=1}^K\frac{L_c}{f_{0c}}\cdot2\sqrt{\frac{\log(2K/\delta)}{2m_c}}.
\end{align*}
The constants $L_c$ are part of the conclusion. No lower bound for this unnormalized total is
claimed. \hfill$\square$

\subsection*{Lower bounds for the label complexity theorem}
The formal lower-bound statements retain the existing labels used elsewhere in the paper.

\begin{lemma}[Necessity, for a suitable shift instance]
\label{supp:lem:necessity}
Fix $\alpha\in(0,1)$. There are a score, a source law, a target covariate law, and constants
$c_0,\varepsilon_0>0$ such that, for every $0<\varepsilon\le\varepsilon_0$, two target laws share
the same $(P_S,P_T(X))$ and satisfy the following property. Every classwise threshold procedure
using the full source law, an unbounded unlabeled target pool, and
$m\le c_0\varepsilon^{-2}$ labeled target pairs has
\[
\max_{P\in\{P_+,P_-\}}
P_{\mathcal A,P}\{\mathcal V(P)^c\cup\mathcal E_\varepsilon(P)^c\}
\ge\frac14.
\]
\end{lemma}

\begin{lemma}[The $\log K$ term is necessary]
\label{supp:lem:logK}
There is $\varepsilon_1>0$ such that for every $0<\varepsilon\le\varepsilon_1$ and every $K\ge4$
there is an instance with $d=\lfloor K/2\rfloor$ informative class pairs, plus one neutral class
when $K$ is odd. Let $m_{j,+}$ and $m_{j,-}$ be fixed, nonadaptive numbers of complete labeled
target pairs sampled from the two classes in pair $j$, and put $n_j=m_{j,+}+m_{j,-}$. When a
neutral class is present, let $m_0$ be its fixed count. If a
classwise-threshold procedure satisfies
\[
\inf_a P_{\mathcal A,P_a}
\{\mathcal V(P_a)\cap\mathcal E_\varepsilon(P_a)\}\ge\frac34
\]
over the full $d$-coordinate hypercube, then, for a constant $C_\alpha>0$ depending only on the
fixed level $\alpha$,
\[
\sum_{j=1}^d e^{-C_\alpha\varepsilon^2 n_j}\le4\log\frac43.
\]
Hence
\[
\sum_{c=1}^{K}m_c=\Omega(K\varepsilon^{-2}\log K),
\]
where a neutral-class count, when present, is included on the left but carries no information.
Equal class counts require $m_c=\Omega(\varepsilon^{-2}\log K)$.
\end{lemma}

\begin{lemma}[Mixed-stream acquisition is necessary]
\label{supp:lem:stream-lower}
For every $d\ge3$, there is a separate instance with $d$ foreground classes, each with probability
$1/(2d)$, and one background class. Suppose a classwise-threshold procedure receives $N$
independent labeled target pairs, as well as the full source law and unlimited unlabeled target
inputs. If
\[
\inf_a P_{\mathcal A,P_a}
\{\mathcal V(P_a)\cap\mathcal E_\varepsilon(P_a)\}\ge\frac34
\]
over the $d$-coordinate hypercube, then
\[
N\ge
\frac{3d}{8c_1^2\varepsilon^2}
\log\!\left\{\frac{d}{4\log 2}\right\},
\]
where the constant $c_1$ is fixed in the construction. In particular,
$N=\Omega(d\varepsilon^{-2}\log d)$.
\end{lemma}

\subsection*{(c) Necessity by a perturbative two point lower bound}
\noindent This is an instance of the classical two point (Le Cam) method
\citep{lecam1973,tsybakov2009}.

\emph{Construction.} Let $X\sim\mathrm{Unif}[0,1]$ under every target law and take
$s(x,c)=x$. For $a\in(-1,1)$ define
\begin{align*}
P_a(Y=c\mid X=x)&=\frac12+a\left(\frac12-x\right),\\
P_a(Y=0\mid X=x)&=\frac12-a\left(\frac12-x\right).
\end{align*}
Both label probabilities lie strictly between zero and one. Also $P_a(Y=c)=1/2$, and the
class-$c$ conditional score CDF is
\begin{align}
F_a(t)
&=P_a(X\le t\mid Y=c)\notag\\
&=2\int_0^t\left\{\frac12+a\left(\frac12-x\right)\right\}\,dx\notag\\
&=t+a(t-t^2),\qquad 0\le t\le1.
\label{eq:two_point_cdf}
\end{align}
Its derivative is $1+a(1-2t)\ge1-|a|>0$. Thus its $(1-\alpha)$-quantile $q(a)$ is unique and solves
\[
q(a)+a\{q(a)-q(a)^2\}=1-\alpha.
\]
Implicit differentiation yields
\begin{equation}
q'(a)=-\frac{q(a)-q(a)^2}{1+a\{1-2q(a)\}}.
\label{eq:q_derivative}
\end{equation}
In particular $q'(0)=-\alpha(1-\alpha)$. By continuity, there are constants
$a_0,d_\alpha>0$ such that
\[
-q'(a)\ge d_\alpha
\quad\text{for every }|a|\le a_0.
\]

Choose a constant $c_1>1/d_\alpha$, set $\tau=c_1\varepsilon$, and restrict
$\varepsilon\le a_0/c_1$. Compare $P_+=P_{+\tau}$ and $P_-=P_{-\tau}$. They have the same
$P_T(X)$; take the source law to be the $a=0$ law under both hypotheses. The unbounded unlabeled
pool and the source therefore have identical distributions. The quantile separation satisfies
\begin{align}
\Delta
&=q(-\tau)-q(+\tau)\notag\\
&=\int_{-\tau}^{+\tau}\{-q'(a)\}\,da\notag\\
&\ge2d_\alpha\tau
>2\varepsilon.
\label{eq:two_point_separation}
\end{align}

\emph{Reduction in full.} Let a threshold procedure output $\hat q_c$. Under $P_a$, its class $c$
part of the common success event is
\[
A_a=
\left\{
F_a(\hat q_c)\ge1-\alpha,\ 
|\hat q_c-q(a)|\le\varepsilon
\right\}.
\]
Since $F_a$ is strictly increasing and $F_a(q(a))=1-\alpha$, validity gives
$\hat q_c\ge q(a)$. Hence
\begin{equation}
A_a\quad\Longrightarrow\quad
q(a)\le\hat q_c\le q(a)+\varepsilon.
\label{eq:two_point_interval}
\end{equation}
Define the test
\[
\widehat a=+\tau
\quad\Longleftrightarrow\quad
\hat q_c<\frac{q(+\tau)+q(-\tau)}2.
\]
Under $P_+$, equations \eqref{eq:two_point_separation} and \eqref{eq:two_point_interval} give
\[
\hat q_c\le q(+\tau)+\varepsilon
<q(+\tau)+\frac{\Delta}{2}
=\frac{q(+\tau)+q(-\tau)}2,
\]
so the test declares $+\tau$. Under $P_-$, validity gives
\[
\hat q_c\ge q(-\tau)
>\frac{q(+\tau)+q(-\tau)}2,
\]
so the test declares $-\tau$. The test can err under $P_a$ only on $A_a^c$. Since the theorem's
simultaneous event implies $A_a$, success probability at least $3/4$ under both hypotheses would
produce a test with each error probability at most $1/4$.

\emph{Information bound.} Write $v(x)=1/2-x$. Conditional on $X=x$, the two hypotheses are
Bernoulli with
\[
p_+(x)=\frac12+\tau v(x),
\qquad
p_-(x)=\frac12-\tau v(x).
\]
For $\tau\le1/2$, both probabilities lie in $[1/4,3/4]$. The elementary Bernoulli bound
$\mathrm{kl}(p,q)\le(p-q)^2/\{q(1-q)\}$ gives
\begin{align*}
\mathrm{KL}(P_+\|P_-)
&\le\frac{64}{3}\tau^2\int_0^1v(x)^2\,dx\\
&=\frac{16}{9}\tau^2.
\end{align*}
For $m$ i.i.d.\ labeled audit pairs, Pinsker's inequality and tensorization \citep{tsybakov2009} imply
\[
\mathrm{TV}(P_+^{\otimes m},P_-^{\otimes m})
\le\sqrt{\frac{m}{2}\mathrm{KL}(P_+\|P_-)}
\le\sqrt{\frac{8m\tau^2}{9}}.
\]
If
\[
m\le\frac{9}{128c_1^2}\varepsilon^{-2},
\]
then this total variation distance is at most $1/4$. The source and unlabeled observations are
identical under the hypotheses, so they add zero total variation. Every test must consequently
have combined error at least
\[
1-\mathrm{TV}(P_+^{\otimes m},P_-^{\otimes m})\ge\frac34.
\]
This contradicts combined error at most $1/2$. In fact one law has failure probability at least
$3/8$, which proves the stated $1/4$ lower bound with
$c_0=9/(128c_1^2)$. \hfill$\square$

\subsection*{(d) Simultaneous necessity and the $\log K$ lower bound}
\noindent This combines the two point method with a product, Assouad style construction across
coordinates \citep{yu1997lecam,tsybakov2009}.

\emph{Construction with complete labeled pairs.} Put $d=\lfloor K/2\rfloor$ and form paired
labels $(j,+)$ and $(j,-)$ for $j=1,\ldots,d$. If $K$ is odd, add one neutral label $0$.
Let $X=(Z,U)$, where $U\sim\mathrm{Unif}[0,1]$. When $K=2d$, take $Z$ uniform on
$\{1,\ldots,d\}$. When $K=2d+1$, set
\[
P(Z=0)=\frac13,
\qquad
P(Z=j)=\frac{2}{3d},\quad j=1,\ldots,d.
\]
The variable $U$ is independent of $Z$, and both are observed. Write $w_j=P(Z=j)$.
For $a=(a_1,\ldots,a_d)\in\{-\tau,+\tau\}^d$, define, in region $Z=j$,
\begin{align*}
P_a\{Y=(j,+)\mid Z=j,U=u\}
  &=\frac12+a_j\left(\frac12-u\right),\\
P_a\{Y=(j,-)\mid Z=j,U=u\}
  &=\frac12-a_j\left(\frac12-u\right).
\end{align*}
When the neutral region is present, set $Y=0$ whenever $Z=0$. For either member
$\sigma\in\{+,-\}$ of pair $j$, let
\[
s((z,u),(j,\sigma))=
\begin{cases}
u,&z=j,\\
M,&z\ne j,
\end{cases}
\qquad M>1,
\]
and set the neutral score to a fixed constant. Take the source law to be the $a=0$ law. Every
hypothesis has the same target covariate marginal and the same source law, and every class has
positive probability.

Conditional on the positive member of pair $j$, the score density is
$f_{a_j}(u)=1+a_j(1-2u)$; conditional on the negative member, it is $f_{-a_j}$. Therefore
\begin{align}
F_{j,+}(t)&=F_{a_j}(t)=t+a_j(t-t^2),\notag\\
F_{j,-}(t)&=F_{-a_j}(t)=t-a_j(t-t^2),
\qquad 0\le t\le1.
\label{eq:hypercube_conditional_cdf}
\end{align}
Their oracle thresholds are $q(a_j)$ and $q(-a_j)$, using the same function as in part~(c).
For $a\ne0$ it is explicitly
\[
q(a)=
\frac{1+a-\sqrt{(1+a)^2-4a(1-\alpha)}}{2a},
\qquad
q(0)=1-\alpha.
\]

For either paired class, the marginal score CDF equals $w_jt$ on $0\le t\le1$. If
\[
\Delta=q(-\tau)-q(+\tau)>0,
\]
then flipping $a_j$ swaps two oracle thresholds separated by $\Delta$, while the pair's total
expected-size contribution is unchanged:
\begin{equation}
w_j\{q(+\tau)+q(-\tau)\}
-w_j\{q(-\tau)+q(+\tau)\}=0.
\label{eq:hypercube_size_gap}
\end{equation}
Thus this construction isolates simultaneous quantile recovery without obtaining a lower bound for
unnormalized total expected set size.

\emph{Reduction in full.} Choose $\tau=c_1\varepsilon$ with $c_1$ fixed so that, for all
sufficiently small $\varepsilon$,
\[
\Delta=q(-\tau)-q(+\tau)>2\varepsilon.
\]
Such a choice exists because
$\Delta=2\alpha(1-\alpha)\tau+O(\tau^3)$. On the event
$\mathcal V(P_a)\cap\mathcal E_\varepsilon(P_a)$, strict monotonicity of the positive member's CDF
turns accuracy and validity into the interval
\begin{equation}
q(a_j)\le\hat q_{j,+}\le q(a_j)+\varepsilon,
\qquad j=1,\ldots,d.
\label{eq:hypercube_efficiency_interval}
\end{equation}
Define
\[
\widehat a_j=+\tau
\quad\Longleftrightarrow\quad
\hat q_{j,+}<\frac{q(+\tau)+q(-\tau)}2.
\]
If $a_j=+\tau$, then \eqref{eq:hypercube_efficiency_interval} and $\varepsilon<\Delta/2$ give
\[
\hat q_{j,+}\le q(+\tau)+\varepsilon
<\frac{q(+\tau)+q(-\tau)}2,
\]
so $\widehat a_j=+\tau$. If $a_j=-\tau$, validity in
\eqref{eq:hypercube_efficiency_interval} gives
\[
\hat q_{j,+}\ge q(-\tau)
>\frac{q(+\tau)+q(-\tau)}2,
\]
so $\widehat a_j=-\tau$. Therefore
\[
\mathcal V(P_a)\cap\mathcal E_\varepsilon(P_a)
\subseteq
\{\widehat a_j=a_j\text{ for every }j\}.
\]
Thus simultaneous validity and threshold accuracy determine the sign vector.

\emph{Product testing experiment.} Give the procedure complete labeled pairs, with fixed counts
$m_{j,+}$ and $m_{j,-}$ from the two classes in pair $j$. Within a stratum, the class label and
region are fixed, and the observed $U$ has density $f_{a_j}$ or $f_{-a_j}$ as above. Thus the full
labeled pair carries exactly the information used below. Different class strata are sampled
independently. The Hellinger affinity
\citep{tsybakov2009} of $f_{+\tau}$ and $f_{-\tau}$ is
\begin{align*}
\rho_1
&=\int_0^1\sqrt{1-4\tau^2(1/2-u)^2}\,du\\
&=1-\eta.
\end{align*}
For $\tau\le1$, the inequalities
$x/2\le1-\sqrt{1-x}\le x$ for $0\le x\le1$ yield
\[
\frac{\tau^2}{6}\le\eta\le\frac{\tau^2}{3}.
\]
Put $n_j=m_{j,+}+m_{j,-}$. Flipping $a_j$ swaps the two densities in both strata, so the affinity
of the complete coordinate experiment is $\rho_{n_j}=(1-\eta)^{n_j}$. Samples from a neutral
class, when present, have a law independent of $a$ and add no information.

Place the uniform product prior on $a\in\{-\tau,+\tau\}^d$. The likelihood and prior both factor
over coordinates, so the posterior factorizes. The Bayes-optimal estimator for exact vector
recovery is coordinatewise MAP, and its all-correct probability is
\[
\prod_{j=1}^d(1-p_j^\star),
\]
where $p_j^\star$ is the equal-prior Bayes error for the complete pair-$j$ experiment. No vector
estimator has larger prior-average exact-recovery probability. A worst-case success probability of
$3/4$ for simultaneous quantile recovery therefore implies
\begin{equation}
\prod_{j=1}^d(1-p_j^\star)\ge\frac34.
\label{eq:product_success}
\end{equation}

The affinity and total variation inequality give
\begin{align*}
p_j^\star
&\ge\frac{1-\sqrt{1-\rho_{n_j}^2}}2\\
&\ge\frac{\rho_{n_j}^2}{4}\\
&=\frac14(1-\eta)^{2n_j}.
\end{align*}
Since $\eta\le\tau^2/3\le1/2$ and $\log(1-\eta)\ge-2\eta$,
\[
p_j^\star
\ge\frac14e^{-4\eta n_j}
\ge\frac14e^{-C_\alpha\varepsilon^2 n_j},
\qquad
C_\alpha=\frac{4c_1^2}{3}.
\]
Taking logarithms in \eqref{eq:product_success} and using
$\log(1-x)\le-x$ gives
\[
\sum_{j=1}^dp_j^\star\le\log\frac43.
\]
It follows that
\begin{equation}
\sum_{j=1}^d e^{-C_\alpha\varepsilon^2n_j}
\le4\log\frac43=:A.
\label{eq:exponential_budget_constraint}
\end{equation}

\emph{Extracting the $\log d$ rate.} Put
$y_j=e^{-C_\alpha\varepsilon^2n_j}$. Jensen's inequality for the convex function
$-\log y$ and \eqref{eq:exponential_budget_constraint} give
\begin{align*}
\sum_{j=1}^d n_j
&=\frac1{C_\alpha\varepsilon^2}\sum_{j=1}^d\log\frac1{y_j}\\
&\ge\frac{d}{C_\alpha\varepsilon^2}
\log\frac{d}{\sum_jy_j}\\
&\ge\frac{d}{C_\alpha\varepsilon^2}\log\frac dA
=\Omega(d\varepsilon^{-2}\log d).
\end{align*}
The total number of stratified labeled pairs is at least $\sum_jn_j$, so with
$d=\lfloor K/2\rfloor$ this is $\Omega(K\varepsilon^{-2}\log K)$. Under equal class counts
$m_{j,+}=m_{j,-}=m$, equation~\eqref{eq:exponential_budget_constraint} directly gives
\[
m\ge\frac1{2C_\alpha\varepsilon^2}\log\frac dA
=\Omega(\varepsilon^{-2}\log K).
\]
This proves Theorem~\ref{thm:labelcomplexity}(d) for complete class-stratified labeled pairs.
\hfill$\square$

\subsection*{Mixed-stream lower bound}
We prove Lemma~\ref{supp:lem:stream-lower} with a separate foreground--background family that gives
the learner every complete labeled pair. Let $X=(Z,U)$ with $Z$ uniform on
$\{1,\ldots,d\}$, $U\sim\mathrm{Unif}[0,1]$, and $Z\perp U$. The labels are
$\{0,1,\ldots,d\}$. In region $Z=c$, set
\begin{align*}
P_a(Y=c\mid Z=c,U=u)&=\frac12+a_c\left(\frac12-u\right),\\
P_a(Y=0\mid Z=c,U=u)&=\frac12-a_c\left(\frac12-u\right),
\end{align*}
with $a_c\in\{-\tau,+\tau\}$. For foreground class $c$, use score $u$ in region $c$ and $M>1$
outside it; use a fixed score for the background. The source is the $a=0$ law. Every hypothesis
has the same target feature law and source law, each foreground class has probability $1/(2d)$,
and its true-class score CDF is $F_{a_c}(t)=t+a_c(t-t^2)$. Thus the coordinatewise interval and
sign test from part~(c) apply. Unlike the fixed-stratified proof, the calculation below retains
background observations and all of their information.

Under every hypothesis, a labeled pair first selects a region $Z=c$ uniformly from
$\{1,\ldots,d\}$. Conditional on this region, its density with respect to uniform $U$ and counting
measure on $Y\in\{0,c\}$ is
\[
g_{a_c}(u,y)=
\begin{cases}
\frac12+a_c(\frac12-u),&y=c,\\
\frac12-a_c(\frac12-u),&y=0.
\end{cases}
\]
\par\noindent
The Hellinger affinity between $g_{+\tau}$ and $g_{-\tau}$ is
\begin{align*}
\rho_1
&=\int_0^1\!\left[
\sqrt{\left\{\frac12+\tau(\frac12-u)\right\}
             \left\{\frac12-\tau(\frac12-u)\right\}}
\right.\\[-2pt]
&\hspace{64pt}\left.
+\sqrt{\left\{\frac12-\tau(\frac12-u)\right\}
             \left\{\frac12+\tau(\frac12-u)\right\}}
\right]du\\
&=\int_0^1\sqrt{1-4\tau^2(1/2-u)^2}\,du
=1-\eta,
\end{align*}
with $\eta\le\tau^2/3$ by the calculation in the fixed-count proof.

First Poissonize the stream. Let the total number of labeled pairs be Poisson with mean $\lambda$.
Poisson splitting makes the $d$ region-specific experiments independent, with each region count
distributed as $\operatorname{Poisson}(\lambda/d)$. For one coordinate, the affinity of the two
resulting compound experiments is
\[
\bar\rho
=\sum_{m=0}^{\infty}
e^{-\lambda/d}\frac{(\lambda/d)^m}{m!}\rho_1^m
=\exp\{-\lambda\eta/d\}.
\]
Thus its equal-prior Bayes error $\bar p^\star$ satisfies
\begin{equation}
\bar p^\star
\ge\frac{\bar\rho^2}{4}
\ge\frac14\exp\!\left\{-\frac{2\lambda\tau^2}{3d}\right\}.
\label{eq:stream_coordinate_error}
\end{equation}
With the uniform product prior on the sign vector, the split experiments and the posterior factor
over coordinates. The best prior-average probability of recovering the complete vector is therefore
$(1-\bar p^\star)^d$, which is at most $\exp(-d\bar p^\star)$.

It remains to return to a fixed stream size. Suppose a procedure using exactly $N\ge1$ labeled
pairs succeeds with probability at least $3/4$ under every sign vector. Run a Poisson experiment
with mean $\lambda=4N$. When it contains at least $N$ pairs, retain the first $N$ and run the fixed
sample procedure; otherwise return any vector. The Poisson count is independent of the labeled
pairs, and a multiplicative Chernoff bound gives
\[
P\{\operatorname{Poisson}(4N)<N\}
\le e^{-9N/8}\le e^{-9/8}.
\]
The resulting prior-average exact-recovery probability is therefore greater than $1/2$. The
Bayes-optimal Poisson procedure must do at least as well, so
\[
\frac12
\le(1-\bar p^\star)^d
\le e^{-d\bar p^\star}.
\]
Combining this inequality with \eqref{eq:stream_coordinate_error} yields
\[
\frac d4\exp\!\left\{-\frac{2\lambda\tau^2}{3d}\right\}
\le\log 2.
\]
Substituting $\lambda=4N$ and $\tau=c_1\varepsilon$ proves
\[
N\ge
\frac{3d}{8c_1^2\varepsilon^2}
\log\!\left\{\frac{d}{4\log 2}\right\}.
\]
Since success in simultaneous validity and threshold recovery determines the complete sign vector
by applying \eqref{eq:two_point_interval} in every region, this proves the lemma. Taking $d=K-1$ proves
Corollary~\ref{cor:stream-lower}. \hfill$\square$

\subsection*{Probability scale quantile recovery}
The efficiency target of the theorem is written in threshold units. The same event carries a
statement in probability units, which needs no density floor and survives a monotone
reparametrization of the score.

\begin{corollary}[Probability scale quantile recovery]
\label{cor:probscale}
Let $\varphi$ be strictly increasing on the range of the score and put $s'=\varphi\circ s$.
\begin{enumerate}
\item[(i)] \emph{Invariance.} PAC Audit Mondrian run on $s'$ returns $\hat q_c'=\varphi(\hat q_c)$,
its prediction sets coincide with those of the run on $s$, and the probability scale excess
$F_c^T(\hat q_c)-(1-\alpha)$ is unchanged, where $\varphi$ is extended by
$\varphi(+\infty)=+\infty$. The threshold error $\hat q_c-q_c^\star$ is not invariant.
\item[(ii)] \emph{Upper bound with no density floor.} Assume only that $F_c^T$ is continuous and
that $(1-\alpha)^{m_c}\le\delta_c$. On the event of part (b), simultaneously for every class,
\[
0\le F_c^T(\hat q_c)-(1-\alpha)\le\eta_c\le2e_c ,
\]
so $m_c=O(\eta^{-2}\log(K/\delta))$ per class suffices for a simultaneous probability scale excess
of at most $\eta$. The hypotheses $f_{0c}>0$, $r_c$ and $B_c\le r_c$ are unused.
\item[(iii)] \emph{The lower bounds transfer.} On the hard families of parts (c) and (d) the
conclusions of Lemmas~\ref{supp:lem:necessity} and \ref{supp:lem:logK} hold with the threshold
criterion $|\hat q_c-q_c^\star|\le\varepsilon$ replaced by the probability scale criterion
$F_c^T(\hat q_c)-(1-\alpha)\le\eta$, with every constant $c_1>1/\{2\alpha(1-\alpha)\}$
admissible once $\eta$ is small.
\end{enumerate}
\end{corollary}

\emph{Proof.} (i) For every $t$ in the range of $s'$,
\[
F_c'^T(t)
=P_T\{\varphi(s(X,c))\le t\mid Y=c\}
=F_c^T(\varphi^{-1}(t)),
\]
which uses strict monotonicity alone. A strictly increasing map preserves the sort order and the
tie pattern of the audit scores, and $k_c$ depends on $m_c$, $\alpha$ and $\delta_c$ alone, so the
procedure selects the same index and $\hat q_c'=\varphi(\hat q_c)$. Hence
$F_c'^T(\hat q_c')=F_c^T(\hat q_c)$, the excess is unchanged, and
\[
\{c:\varphi(s(x,c))\le\varphi(\hat q_c)\}=\{c:s(x,c)\le\hat q_c\},
\]
so the prediction set is unchanged as well. The argument allows a different $\varphi_c$ in each
class, and the convention $\varphi(+\infty)=+\infty$ covers the classes whose threshold is
infinite. For a $\varphi$ that is continuous and differentiable near $q_c^\star$, the oracle
quantile transforms as $q_c'^\star=\varphi(q_c^\star)$ and
\[
\varphi(\hat q_c)-\varphi(q_c^\star)
=\varphi'(q_c^\star)(\hat q_c-q_c^\star)+o(|\hat q_c-q_c^\star|)
\]
as $\hat q_c$ tends to $q_c^\star$, so a non constant derivative generally changes the threshold
error. For a $\varphi$ with jumps the transformed oracle quantile need not be
$\varphi(q_c^\star)$ at all.

(ii) These are the two inequalities already proved, the lower one in part (a) and the upper one in
\eqref{eq:pac_mondrian_cdf_upper}, read before the density floor is used. The floor $f_{0c}$ and
the interval length $r_c$ enter the proof of part (b) only where a CDF error is converted into a
threshold error. Inverting \eqref{eq:eta_hoeffding} for $2e_c\le\eta$ gives the stated sample size.

(iii) All lower-bound families use the class-conditional CDFs $F_a(t)=t+a(t-t^2)$ of
\eqref{eq:two_point_cdf} with $a=\pm\tau$. The fixed-stratified construction places $F_a$ and
$F_{-a}$ in each class pair, while the mixed stream places one $F_a$ in each foreground region.
At a common threshold
$t$ the two laws separate in CDF value by
\[
F_{+\tau}(t)-F_{-\tau}(t)=2\tau(t-t^2),
\]
which at $t$ near $q(0)=1-\alpha$ equals $2\tau\alpha(1-\alpha)+O(\tau^2)$ and is therefore of the
same order in $\tau$ as the quantile separation \eqref{eq:two_point_separation}. Under $P_a$,
validity together with a probability scale excess of at most $\eta$ gives
\[
q(a)\le\hat q_c\le F_a^{-1}(1-\alpha+\eta),
\]
the analogue of \eqref{eq:two_point_interval}. Applying the increasing map $F_{+\tau}$ to both
sides and using $F_{-\tau}(q(-\tau))=1-\alpha$, these two intervals are disjoint once
\[
\eta<2\tau\left\{q(-\tau)-q(-\tau)^2\right\},
\]
whose right side tends to $2\tau\alpha(1-\alpha)$ as $\tau$ tends to zero. Choosing
$\tau=c_1\eta$ with $c_1>1/\{2\alpha(1-\alpha)\}$ therefore separates the two laws for small
$\eta$. This constant sits below the $1/d_\alpha$ of Lemma~\ref{supp:lem:necessity} only because
the test used here is one sided. Validity pins $\hat q_c$ above $q(a)$ on the threshold scale as
well, where the same one sided test admits every $c_1>1/(2d_\alpha)$, so the change of units
leaves the constant unchanged. The test
becomes the coordinatewise rule that declares $+\tau$ exactly when $\hat q_c<q(-\tau)$, which is
correct on the success event under both laws by the displayed interval. The information bound of
parts (c) and (d) is then used word for word, since it depends on $\tau$ and the sample size
alone. The transfer is free because this family has a score
density bounded above and below by absolute constants near the quantile, which is the
bi-Lipschitz condition Remark~\ref{rem:whynotsize} identifies as what a change of units requires.
\hfill$\square$

\begin{remark}[The two scales are not equivalent in general]
\label{rem:probscale_scope}
Part (iii) is a property of this hard family. It is not a general equivalence of the two criteria.
On a family whose score density vanishes at the quantile, a small probability scale excess is
compatible with a large threshold error, and the two criteria come apart. The quantity the
expected set size corollary bounds, $\E_{P_T}|C_{\hat q}(X)|-\sum_cG_c(q_c^\star)$, is
invariant by the set invariance of part (i). Its bound $\sum_cL_cB_c$ is not. A Lipschitz constant
and a density floor are a supremum and an infimum taken over an interval, and they need not be
attained at the same point, so their ratio moves under a reparametrization.
\end{remark}

\begin{remark}[Scope of the theorem]
Validity is the single PAC event $\mathcal V(P)$ in all four parts. Part (b) uses a local density
lower bound to convert CDF error into threshold error. Part (c) uses independent labeled target
pairs; part (d) allows complete labeled pairs under fixed, nonadaptive class-stratified counts. The mixed-stream lower bound
also uses independent target pairs and does not cover adaptive label acquisition. The matching
statements concern class-conditional quantile recovery. The expected-size result is the one-sided
corollary above, with each marginal sensitivity $L_c$ visible.
\end{remark}

\begin{remark}[Why the estimand is quantile recovery and not expected set size]
\label{rem:whynotsize}
The construction in part~(d) explains why the matching lower bound concerns quantile recovery.
In the paired hard family, changing one coordinate swaps two class thresholds separated by
$\Delta>2\varepsilon$ but leaves their combined expected-size contribution unchanged; see
\eqref{eq:hypercube_size_gap}. This cancellation keeps the $d$ coordinates available for product
testing, but it prevents the reduction from giving a matching bound for total expected set size.
On any segment where $G_c$ is quantitatively invertible, a per-class size criterion is
equivalent to threshold recovery up to its slope. The corollary above gives the one-sided relation
available for the total size.
\end{remark}

%% file: sections/alt_appendix_experiments.tex
\section{Experimental details}
\label{app:experiments}

\subsection{Evaluation measures}
For each method, marginal coverage is the fraction of test examples whose prediction set contains
the true label. Class coverage is computed separately for every class. Worst-class coverage is
their minimum. For $K$ classes, put $r_K=\max\{1,\lceil0.1K\rceil\}$; then
$\mathrm{CVaR}_{10}$ is the mean of the $r_K$ smallest class coverages. Mean set size is computed
on the same target evaluation examples. For Table~\ref{tab:alt-real-shifts}, each class coverage is
first averaged over ten source/calibration seeds, after which the minimum and lower-tail mean are
computed. We call this the aggregate coverage profile. The paired uncertainty summary below instead
forms a method difference within each seed.
The ten-seed studies use integer split seeds $0,\ldots,9$.

\subsection{Real-shift protocols}
The NTU RGB+D 60 \citep{shahroudy2016ntu} cross-subject and cross-view studies use a compact
ST-GCN \citep{yan2018stgcn} trained for 80 epochs from scratch on the source training split.
Per-frame features are summarized by a 64-dimensional
regularized SPD covariance descriptor, mapped to a tangent space, and classified by multinomial
logistic regression. Backbone training, classifier fitting, and calibration use disjoint
sequences. Each class has about 334 source calibration examples. NTU to N-UCLA
\citep{wang2014nucla} uses raw-skeleton SPD descriptors with dimension 120, mean-spine
normalization, the 20 shared joints, and the five shared actions; this avoids assuming a feature
extractor shared by the two sensors.

For each integer seed, the source cache is split into 40\% scorer fitting, 35\% calibration, and
25\% held-out data. The target cache is split equally into an unlabeled pool and evaluation data.
Zero-shot image scorers require no source fitting, but retain the same 35\% source calibration and
50\% target evaluation fractions so that the method comparisons use the common split protocol.

The target-label curve in Figure~\ref{fig:alt-budget} uses a separate class-stratified protocol.
Within each target class, 40\% of the examples are reserved for evaluation and the remaining 60\%
form the audit pool. Each plotted budget is drawn without replacement from that pool; the smallest
class contains 161 available audit examples. The same audit and evaluation indices are used by
the paired methods within a resample.

The image studies use probabilities from fixed zero-shot CLIP classifiers \citep{radford2021clip}
and the prompt ``a photo of a \{class name\}.'' CIFAR-100 \citep{krizhevsky2009cifar} uses 50,000 clean training images as the source and
10,000 test images with independently generated Gaussian pixel noise of scale $\sigma=0.03$ as the
target. Its checkpoint is OpenAI ViT-L-14-quickgelu. Office-Home and DomainNet use
ViT-B/32 with LAION-2B-s34B-b79K weights on the Clipart-to-Real and Sketch-to-Real domain shifts
\citep{venkateswara2017officehome,peng2019domainnet};
their target domains contain 4,357 and 175,327 images. Each checkpoint uses its standard
OpenCLIP preprocessing and learned logit scale, with no added temperature. The CIFAR shift is a
controlled corruption and not the distributed CIFAR-100-C benchmark.
The nonconformity score is $s(x,c)=1-\hat p(c\mid x)$ in the main results. Within each study, every
calibration rule uses the same scorer, split, and evaluation examples.

The robustness figure uses CTR-GCN \citep{chen2021ctrgcn} and compares LAC
\citep{sadinle2019lac}, APS \citep{romano2020aps}, and RAPS \citep{angelopoulos2021raps}. If
$r_x(c)$ is the rank of class $c$ under
$\hat p(\cdot\mid x)$ and $U\sim\mathrm{Unif}[0,1]$ is shared by the paired methods, the randomized
APS score is
\[
  s_{\mathrm{APS}}(x,c)=
  \sum_{j:\,\hat p(j\mid x)>\hat p(c\mid x)}\hat p(j\mid x)+U\hat p(c\mid x).
\]
RAPS adds $\lambda\max\{r_x(c)-k_{\mathrm{reg}},0\}$ with
$\lambda=0.01$ and $k_{\mathrm{reg}}=5$. The principal tables use LAC only, so these score choices
do not change their estimand.

Split CP uses one source quantile; Source Mondrian uses one per class. Weighted CP forms its source
quantile with source-to-target marginal density ratios from an out-of-sample,
$\ell_2$-regularized logistic domain classifier with balanced domain classes. Ratios are clipped to
$[0,20]$, and the weighted quantile includes the standard test atom at $+\infty$. The target oracle
uses disjoint target calibration and evaluation splits. Target labels used for calibration,
matching, or evaluation are excluded from label-free methods. These common-scorer comparisons are
diagnostic, not a claim of a complete state-of-the-art benchmark.

\subsection{Experimental settings}
Table~\ref{tab:alt-repro} collects the settings shared across the reported studies.

\begin{table}[htbp]
\centering
\footnotesize
\setlength{\tabcolsep}{4pt}
\renewcommand{\arraystretch}{1.02}
\caption{Core experimental and controlled-study settings.}
\label{tab:alt-repro}
\begin{tabular*}{\linewidth}{@{\extracolsep{\fill}}L{.21\linewidth}L{.73\linewidth}@{}}
\toprule
Component & Setting \\
\midrule
Skeleton backbone & 80 epochs; Adam; learning rate $0.01$; cosine schedule; weight decay
$10^{-4}$; batch size 32; 64 resampled frames \\
SPD scorer & 64-dimensional covariance; diagonal regularization $10^{-6}$; log map;
standardization; PCA to at most 256 components; logistic regression with $C=1$ and 2000 iterations \\
Ratio model & Standardized logistic regression; $C=1$; balanced domain classes; 5000 iterations;
ratio clipped to $[0,20]$ \\
Scaling grid & $K\in\{5,10,30,60,100,300\}$,
$\delta\in\{.20,.10,.05,.01\}$, $\varepsilon\in\{.02,.04,.08\}$, and scale multipliers
$\{.25,.5,1,2,4\}$; 2000 repetitions per grid point \\
Common transition diagnostic & First sample size reaching 90\% joint success, interpolated linearly in log
sample size between adjacent grid points; unresolved and left-censored crossings are reported \\
Random counts & $K=345$, $\alpha=.10$, $\delta=.05$, 1000 simulations; balanced, Zipf with
exponent 1,
$3{:}1$ geometric, and $10{:}1$ geometric priors \\
\bottomrule
\end{tabular*}
\end{table}

\subsection{Controlled checks}
The controlled studies use the fixed settings in Table~\ref{tab:alt-repro} and report every tested
condition, including unresolved or left-censored transitions.

\paragraph{Checks aligned with the theory.}
The coupled identification study evaluates 180 pairs of target laws that agree on every
label-free observation. Outputs are bit-identical within each pair, and the empirical coverage/size
calculation differs from its analytic value by at most $0.0058$. For the exact one-sided rule with
continuous scores, coverage averaged over audits is $k/(m+1)$: $0.9804$, $0.9505$, and $0.9403$ at
$m=50,100,200$. This average measures conservatism; PAC success requires
$F_c^T(\hat q_c)\ge0.90$ for every class. Across all 360 scaling cells, empirical joint validity
exceeded $1-\delta$. The closed-form Beta probability exceeded that target by at least $0.005$, and
the empirical and closed-form rates differed by at most $0.0171$. Known CDFs avoid held-out noise.

The scaling study uses $F_a(t)=t+a(t-t^2)$ and varies $K$, the target accuracy, and confidence.
Of 72 groups, 71 have a resolved common 90\% success transition; normalized sample sizes span
$0.837$--$2.221$ times the theoretical scale. This supports the predicted
$\varepsilon^{-2}\log(K/\delta)$ order, but one fixed factor-two window is too narrow for all finite
settings.

The random-count study fixes $K=345$ and obtains $m_\star=91$ as the finite-threshold class quota.
Across balanced, Zipf, $3{:}1$ geometric, and $10{:}1$ geometric priors, the probability that every
class reaches its quota is at least $0.90$ once $N\pi_{\min}/m_\star=1.5$. At $N=Km_\star$,
deterministic quotas have no infinite thresholds, while mixed-stream sampling under the $10{:}1$
prior leaves an average fraction $0.5897$ infinite. This separates the estimation budget after
stratification from the draw budget needed for rare classes.

\paragraph{Diagnostics with different estimands.}
The controlled record also asked whether every finite-rule coverage lay within $0.03$ of $0.90$;
72 of 144 readings did not. This measures closeness to the boundary, not the one-sided guarantee,
and the exact rule is expected to be conservative. The factor-two scaling window likewise tests a
finite-sample constant, not the order result. We report both diagnostics but do not treat them as
tests of the theorem.

\paragraph{Finite-evaluation diagnostic.}
On NTU-60 cross-subject, the exact order-statistic audit first has a finite threshold for every
class at $m=74$. At the evaluated budgets $m=79,99,149$, its mean set size is respectively
$4.425$, $2.863$, and $2.250$ times that of ordinary Target Mondrian. The corresponding fractions
of audits whose held-out coverage estimate exceeds $0.90$ in every class are $0.910$, $0.810$, and
$0.780$. These fractions estimate a simultaneous minimum from finite held-out samples, not the
theorem's PAC event; binomial noise compounds over 60 classes. The controlled study checks that
event with exact CDFs, while the NTU experiment records finite-threshold and efficiency costs.

\subsection{Paired variation across source splits}
Table~\ref{tab:alt-paired} reports seed-wise Source Mondrian--Split CP differences with descriptive
95\% $t$ intervals. Because the main table ranks after averaging, its nonlinear minimum and
lower-tail differences need not equal the mean seed-wise effects.

\begin{table}[htbp]
\centering
\footnotesize
\setlength{\tabcolsep}{3pt}
\caption{Seed-wise Source Mondrian--Split CP effects over ten source splits; descriptive 95\%
$t$ intervals and positive-difference counts.}
\label{tab:alt-paired}
\begin{tabular*}{\linewidth}{@{\extracolsep{\fill}}L{.19\linewidth}L{.265\linewidth}L{.265\linewidth}cc@{}}
\toprule
Setting & $\Delta\cvar$ [95\% CI] & $\Delta$ worst [95\% CI] & $\cvar$ wins & Worst wins \\
\midrule
NTU subject & $0.057\ [0.042,0.071]$ & $0.062\ [0.037,0.087]$ & 10 & 10 \\
NTU view & $0.053\ [0.040,0.066]$ & $0.060\ [0.037,0.084]$ & 10 & 10 \\
NTU $\to$ N-UCLA & $0.001\ [-0.002,0.005]$ & $0.001\ [-0.002,0.005]$ & 1 & 1 \\
CIFAR-100 corruption & $0.124\ [0.106,0.142]$ & $0.229\ [0.189,0.270]$ & 10 & 10 \\
Office-Home C $\to$ R & $0.049\ [0.022,0.076]$ & $0.574\ [0.490,0.657]$ & 8 & 10 \\
DomainNet S $\to$ R & $-0.098\ [-0.103,-0.093]$ & $0.101\ [0.063,0.138]$ & 0 & 10 \\
\bottomrule
\end{tabular*}
\end{table}
\FloatBarrier

\subsection{Pure label shift as an identifying boundary}
\label{app:label_shift}
Under pure label shift, the source and target laws share $P(X\mid Y=c)$ and differ only in their
class priors. For a fixed scorer this gives, for every class,
\[
  P_T\{s(X,c)\le t\mid Y=c\}=P_S\{s(X,c)\le t\mid Y=c\}.
\]
Thus source calibration scores are exchangeable with a target test score from the same class, and
Proposition~\ref{prop:mondrian} gives unconditional per-class coverage. This equality is an
identifying boundary of the impossibility result, not an empirical claim.

We check a controlled generator with $K\in\{10,60\}$, LAC and APS scores, nominal coverage $0.90$,
and ten replicates, requiring 30 expected evaluation examples per class. Table~\ref{tab:alt-label-shift-all}
uses each replicate's last admissible setting, independent of reaching a requested match; mean
$\kappa$ is descriptive.

\begin{center}
\begin{minipage}{.94\linewidth}
\centering
\footnotesize
\setlength{\tabcolsep}{3.1pt}
\renewcommand{\arraystretch}{0.97}
\refstepcounter{table}\label{tab:alt-label-shift-all}%
\small\textbf{Table~\thetable:}
Pure label-shift check over ten replicates; Ratio is a misspecified marginal-density-ratio control.
\par\vspace{4pt}
\footnotesize
\begin{tabular}{@{}rrrccccr@{}}
\toprule
$K$ & Score & Mean $\kappa$ & Split marg. & Split worst & Source marg. & Source worst & Ratio worst\\
\midrule
10 & LAC & 4.05 & 0.887 & 0.879 & 0.902 & 0.891 & 0.828\\
10 & APS & 4.05 & 0.899 & 0.886 & 0.902 & 0.887 & 0.865\\
60 & LAC & 8.15 & 0.895 & 0.887 & 0.900 & 0.892 & 0.799\\
60 & APS & 8.15 & 0.897 & 0.890 & 0.900 & 0.890 & 0.822\\
\bottomrule
\end{tabular}
\end{minipage}
\end{center}

Source Mondrian stays near nominal because pure label shift preserves each within-class score law.
The marginal-ratio control targets a different relation and undercovers. Pointwise intervals ignore
shared thresholds and multiplicity; Proposition~\ref{prop:mondrian} instead gives unconditional
coverage separately by class. This check is limited to the controlled generator and prior range.
It illustrates the identifying boundary, but does not show that every classifier is insensitive to
label shift or that marginal coverage stays constant as class priors change.